\PassOptionsToPackage{dvipsnames,svgnames}{xcolor}
\documentclass{article} 
\usepackage{iclr2025_conference,times}

\usepackage{amsmath,amsfonts,bm}

\def\eqref#1{equation~\ref{#1}}

\DeclareMathAlphabet{\mathsfit}{\encodingdefault}{\sfdefault}{m}{sl}
\SetMathAlphabet{\mathsfit}{bold}{\encodingdefault}{\sfdefault}{bx}{n}

\newcommand{\seq}{\mathbf{a}}

\def\gR{{\mathcal{R}}}

\def\sI{{\mathbb{I}}}

\newcommand{\E}{\mathbb{E}}

\newcommand{\R}{\mathbb{R}}

\DeclareMathOperator*{\argmin}{arg\,min}

\DeclareMathOperator{\Tr}{Tr}

\usepackage{booktabs}       
\usepackage{nicefrac}       
\usepackage{microtype}      
\usepackage{multirow}
\usepackage{colortbl}
\usepackage{newtxmath}
\usepackage{pdfpages}
\usepackage{dcolumn}
\newcolumntype{d}[1]{D{.}{.}{#1}}
\usepackage{longtable}
\usepackage{threeparttable}
\usepackage{threeparttablex}
\usepackage{enumitem}
\usepackage{caption}
\usepackage{subfigure}
\usepackage{wrapfig}
\usepackage{graphicx}
\usepackage{amsmath,amscd,amsbsy,amsfonts,latexsym,url,bm,bbm,dsfont}
\usepackage{algorithm}
\usepackage{algorithmic}
\usepackage[normalem]{ulem}

\usepackage[algo2e,ruled,vlined,boxed,linesnumbered]{algorithm2e}
\definecolor{mydarkblue}{rgb}{0,0.08,0.45}
\definecolor{mydarkred}{rgb}{0.6,0,0}
\definecolor{myblue}{HTML}{268BD2}
\usepackage[colorlinks,
linkcolor=mydarkred,
citecolor=myblue]{hyperref}
\allowdisplaybreaks[4]

\allowdisplaybreaks

\newcommand{\abs}[1]{\left|#1\right|}

\newtheorem{Lemma}{Lemma}

\newtheorem{Theorem}{Theorem}

\newtheorem{Proof}{Proof}

\newcommand{\Prob}{{\mathbb{P}}}

\newcommand{\D}{{\mathcal{D}}}

\newcommand{\II}{{\mathcal{I}}}

\newcommand{\QQ}{{\mathcal{Q}}}

\newcommand{\highlight}[1]{{\textcolor{MediumSeaGreen}{#1}}}

\newcommand{\firstplace}[1]{\cellcolor[HTML]{FFB3B3}\textbf{#1}} 
\newcommand{\secondplace}[1]{\cellcolor[HTML]{FFCCCC}#1} 
\newcommand{\thirdplace}[1]{\cellcolor[HTML]{FFE5E5}#1} 
\newcommand{\baseline}[1]{\cellcolor[HTML]{CDEBFC}#1} 

\newcommand{\approach}{\textsc{FactTest}\xspace}
\newcommand{\mini}{\textsc{FTest}\xspace}

\title{\approach: Factuality Testing in Large Language Models with Finite-Sample and Distribution-Free Guarantees}

\iclrfinalcopy

\author{Fan Nie$^{1}$ \hspace{0.1cm} Xiaotian Hou$^{2}$ \hspace{0.1cm} Shuhang Lin$^{2}$ \hspace{0.1cm} James Zou$^{1}$ \hspace{0.1cm} Huaxiu Yao$^{3}$ \hspace{0.1cm} Linjun Zhang$^{2}$\thanks{Corresponding author.}\\
$^{1}$Stanford University, $^{2}$Rutgers University, $^{3}$UNC-Chapel Hill\\
}

\begin{document}

\maketitle

\begin{abstract}
The propensity of Large Language Models (LLMs) to generate hallucinations and non-factual content undermines their reliability in high-stakes domains, where rigorous control over Type I errors (the conditional probability of incorrectly classifying hallucinations as truthful content) is essential. Despite its importance, formal verification of LLM factuality with such guarantees remains largely unexplored.
In this paper, we introduce \textbf{\approach}, a novel framework that statistically assesses whether a LLM can confidently provide correct answers to given questions with finite-sample and distribution-free correctness guarantees. We formulate factuality testing as hypothesis testing problem to enforce an upper bound of Type I errors at user-specified significance levels. Notably, we prove that our framework also ensures strong Type II error control under mild conditions and can be extended to maintain its effectiveness when covariate shifts exist. 
Our approach is distribution-free and works for any number of human-annotated samples. It is model-agnostic and applies to any black-box or white-box LM. Extensive experiments on question-answering (QA) and multiple-choice benchmarks demonstrate that \approach effectively detects hallucinations and improves the model's ability to abstain from answering unknown questions, leading to an over 40\% accuracy improvement.
\end{abstract}

\section{Introduction}\label{sec:intro}

Large Language Models (LLMs) like ChatGPT and GPT-4~\citep{ouyang2022training,gpt4} have demonstrated substantial advancements in various domains such as summarization systems, search engines and virtual assistants. However, their outputs cannot be fully trusted due to their propensity to generate nonfactual and incorrect information with seemingly high confidence, a challenge often referred to as hallucination~\citep{maynez2020faithfulness,huang2023survey,ji2023survey}. This tendency undermines the reliability and trustworthiness of the generated content, highlighting a critical need for robust mechanisms to verify the factuality and correctness of LLM outputs.

Existing approaches to hallucination detection like retrieval-based methods~\citep{thorne2018factextractionverificationfever, gou2024criticlargelanguagemodels, chen2024complexclaimverificationevidence} and training-based approaches~\citet{zhang2023r} either rely on external databases or resource-intensive fine-tuning process, which are often impractical or costly. Therefore, there has been growing interest in uncertainty estimation as a zero-resource alternative for hallucination detection~\citep{varshney2023stitch,xiong2024llmsexpressuncertaintyempirical}, operating under the premise that hallucinations are intrinsically tied to the model’s uncertainty~\cite{huang2023survey}. However, none of these methods can provide theoretical guarantees for the detection or testing results, which are essential for deploying LLMs in high-stakes domains~\cite{kumar2023conformal} where precise control of Type I errors (incorrectly flagging a hallucination as
truthful content) is needed for decision-making. For instance, incorrect medical diagnoses in healthcare or the provision of uncertain legal advice in the legal field could result in detrimental consequences.

To address these limitations, we introduce \approach, a framework that statistically evaluates whether an LLM can reliably generate correct answers to given questions with provable correctness guarantees and teach LLMs to abstain from answering uncertain questions. We formulate the factuality testing within a hypothesis testing framework to theoretically control the Type I error while minimizing the Type II error. Leveraging the fundamental connection between Neyman-Pearson (NP) classification and statistical testing \citep{tong2018neyman,JMLR:v14:tong13a,scott2005neyman}, we define a score function to quantify model certainty and select an appropriate threshold based on a constructed calibration dataset to instruct LLMs to refuse uncertain questions and control the false positive rate.
Furthermore, we prove that, under mild conditions, \approach achieves strong power control. This ensures that our method not only controls Type I error but also maintains a low Type II error, thereby providing reliable factuality assessments. On the other hand, since statistical tests often rely on the i.i.d. assumption, which may not hold in reality, we enhance the robustness of this framework by adding an extension to accommodate covariate shifts through the estimation of density ratios and the use of rejection sampling. Our approach is model-agnostic and does not rely on specific data distribution assumptions, making it broadly applicable to any language model. Importantly, it works for any finite number of human-annotated samples, ensuring practicality and ease of implementation. 

To the best of our knowledge, this study is the first to introduce statistical factuality testing in large language models, thereby facilitating safer and more reliable deployment in high-stakes applications. We evaluate the effectiveness of our proposed framework on question-answering (QA) and multiple-choice benchmarks. The results indicate that our approach offers several significant advantages: (1) it consistently outperforms base models by a substantial margin without requiring additional training or external data sources; (2) it surpasses fine-tuned baselines by a large margin while utilizing only half of the training data; and (3) it maintains superior performance on out-of-distribution testing data. Notably, the theoretical guarantees of our method remain valid even when the i.i.d. assumption is violated.
\textbf{We summarize the main contributions below}.

\setlength{\leftmargini}{20pt}
\begin{itemize} 
\item We propose \approach, a novel statistical testing framework that evaluates the factuality of LLMs while teaching them to decline uncertain questions with user-specified Type I error guarantees.
\item We prove that our statistical framework achieves strong power control under mild conditions, ensuring that the predictor can also maintain a low Type II error. This power analysis is directly applicable to the standard NP classification problems, not limited to this setting.
\item We extend our framework to accommodate covariate shifts by approximating density ratios and applying rejection sampling, thereby enhancing its robustness in real-world applications.
\item We demonstrate that \approach effectively detects hallucinations while maintaining Type I error below user-specified significance levels, achieving an over 40\% improvement in accuracy compared to pretrained models without any fine-tuning. Additionally, it surpasses training-based baselines by 30\% using only half of the fine-tuning data.
\end{itemize}

\vspace{0pt}\section{Statistical Factuality Testing}\label{sec:method}\vspace{0pt}

In this section, we formulate the evaluation of factuality in LLMs as a statistical hypothesis testing problem and introduce our \approach framework to overcome hallucination issues.

\subsection{Problem Formulation}\label{sec:formulation}\vspace{0pt}

We consider a text generation task in which a language model $M$ will generate its answers $M(q)$ based on a question $q$. Our goal is to statistically evaluate whether $M$ can correctly answer $q$, termed as $M$ being certain of $q$. We formulate this objective as a hypothesis testing problem with the following hypotheses:
\begin{align*}
    H_0 &: \text{The model } M \text{ cannot answer the question } q \text{ certainly.} \\
    H_1 &: \text{The model } M \text{ can answer the question } q \text{ certainly.}
\end{align*}
For any question-answer pair $(q,a)$ with $a$ to be the correct answer for question $q$, we apply $M$ to generate an answer $M(q)$. Then we classify the question-generated answer pair $(q,M(q))$ as certain if the null hypothesis $H_0$ is rejected, i.e., $M(q)$ aligns with $a$; otherwise, it is deemed uncertain. Let $P_0$ and $P_1$ represent the distributions of uncertain and certain question-generated answer pair $(q,M(q))$, respectively. 

Given a dataset $\{(q_1, a_1), ...,(q_n, a_n)\} \subset \mathcal Q\times\mathcal A$ comprising 
$n$ question-answer pairs, we apply $M$ to generate answers for all questions, resulting in the set $\D=\{(q_1,M(q_1),a_1),\ldots, (q_n,M(q_n),a_n)\}$. Then our goal is to construct a predictor $\hat f_\alpha: \mathcal Q\times\mathcal A \to \{0,1\}$ that classifies a pair $(q,M(q))$ as certain (output 1) or uncertain (output 0) while ensuring that the false positive rate, or Type I error, does not exceed a pre-specified significance level $\alpha$. 
Formally, we seek $\hat f_\alpha$ such that the error of predicting uncertain $(q,M(q))$ as certain is below level $\alpha$ with probability at least $1-\delta$, i.e.,
\begin{equation}\label{eq:target1}
    \Prob_\D(\Prob_{(q,M(q))\sim P_0}(\hat f_\alpha(q,M(q))=1)>\alpha)\le\delta.
\end{equation}
where $\delta$ denotes the allowable probability of exceeding the significance level. Note that given any question $q$, the answer $M(q)$ generated by $M$ is randomized. While the distribution of $M(q)$ is fully determined by $q$, the realization $M(q)$ involves additional sampling randomness independent of $q$. By taking $(q,M(q))$ as inputs to $\hat f_\alpha$, we enable the predictor to utilize information from the question $q$, the distribution of $M(q)$ (by asking $M$ the same question $q$ multiple times), and the current realization $M(q)$ of the produced answer.

\subsection{Finite-sample and Distribution-free Type I Error Control} \label{section:calibration}
\textbf{Calibration Dataset Construction.}
Following the methodology of \citet{zhang2023r}, we adopt a supervised identification strategy to partition the dataset $\D$ into a certain subset $\D_1$ and an uncertain subset $\D_0$. 

Specifically, for each question-generated answer pair $(q_i, M(q_i))$ in $\D$, we define an indicator variable $y_i\in\{0,1\}$ to indicate whether $M$ is certain about $q_i$ such that
\[
y_i =
\begin{cases}
1, & \text{if } M(q_i) \text{ aligns with the true answer } a_i, \\
0, & \text{otherwise}.
\end{cases}
\]
Based on these indicators, the dataset is divided into:
\[
\mathcal{D}_1 = \{ (q_i,M(q_i)) \in \mathcal{Q}\times\mathcal{A} : y_i = 1, i\in[n] \}, \quad \mathcal{D}_0 = \{ (q_i,M(q_i)) \in \mathcal{Q}\times\mathcal{A} : y_i = 0, i\in [n]\}.
\]
Assuming that \(\{(q_i, M(q_i), y_i)\}_{i=1}^n\) are independent and identically distributed (i.i.d.) samples from distribution \( P \), and that the distributions of \(\mathcal{D}_0\) and \(\mathcal{D}_1\) are \( P_0 \) and \( P_1 \), respectively.


\textbf{Certainty Predictor based on Score Function.}
Suppose there is a score function $\hat\eta:\mathcal Q\times\mathcal A\to \R$ that measures the certainty. The value is expected to be large if $M$ has the ability to provide a factual answer. The predictor $\hat f_\alpha(q,M(q))$ can then be defined as:
\begin{equation}\label{eq:function}
\hat f_\alpha(q,M(q))=\sI(\hat \eta(q,M(q))>\hat\tau_\alpha)
\end{equation}
where $\sI$ is the indicator function and $\hat\tau_\alpha$ is a threshold to be determined. The task thus reduces to selecting a threshold $\hat\tau_\alpha$ that satisfies the requirement in Eq.~\ref{eq:target1}:
\begin{equation}\label{eq:target2}
    \Prob_\D(\Prob_{(q,M(q))\sim P_0}(\hat\eta(q,M(q))>\hat\tau_\alpha)>\alpha)\le\delta.
\end{equation}



\textbf{Calibration and Threshold Selection}
To determine the appropriate threshold $\hat\tau_\alpha$, we utilize the calibration subset $\D_0$. Denote the $n_0$ samples in $\D_0$ as $\D_0=\{(q_i^{(0)},M(q_i^{(0)})):i\in[n_0]\}$. For each calibration sample $(q_i^{(0)},M(q_i^{(0)})) \in \D_0$, we compute the certainty score $T_i=\hat\eta(q_i^{(0)},M(q_i^{(0)}))$. We then order these scores in ascending order to obtain the order statistics $T_{(1)}\le \ldots\le T_{(n_0)}$, and set $T_{(n_0+1)}=+\infty$. Motivated by \cite{tong2018neyman}, the threshold $\hat\tau_\alpha$ is selected based on the following probabilistic guarantee:
\begin{equation}\label{eq:tau}
    \Prob_\D(\Prob_{(q,M(q))\sim P_0}(\hat\eta(q,M(q))>T_{(k)})>\alpha)\le\sum_{j=k}^{n_0}{n_0\choose j}(1-\alpha)^j\alpha^{n_0-j}\overset{\triangle}{=}v(k),\quad k\in[n_0+1]
\end{equation}
    when $k=n_0+1$, $v(k)$ is defined to be 0. We then determine $\hat k$ as
\begin{equation}
\hat k=\min\{k\in[n_0+1]:v(k)\le\delta\},
\end{equation}
Subsequently, the threshold is set to: $\hat\tau_\alpha=T_{(\hat k)}$. Note that $\hat\tau_\alpha$ is well defined for any $n_0$, ensuring Type I error control irrespective of the calibration sample size $n$. Specifically, when $n_0$ is small such that $v(n_0)>\delta$, the threshold becomes $\hat\tau_\alpha=T_{(n_0+1)}=+\infty$, causing $\hat f_\alpha$ to conservatively classify all pairs $(q,M(q))$ as uncertain, thereby abstaining from answering any question.
The derivation is deferred to Appendix~\ref{sec:proof}.

\begin{Theorem}\label{thm:type1}
    For any $n\in\mathbb{N}_+$, with probability at least $1-\delta$, the constructed classifier $\hat f_\alpha$ has type I error below $\alpha$, i.e., 
    \vspace{-3pt}
    \[\Prob_\D\big(\Prob_{(q,M(q))\sim P_0}(\hat f_\alpha(q,M(q))=1)\le\alpha\big)\ge 1-\delta.\]
\end{Theorem}

This theorem provides a finite-sample and distribution-free guarantee of Type I error control. With the determined threshold $\hat\tau_\alpha$, the predictor $\hat f_\alpha(q,M(q))=\sI(\hat \eta(q,M(q))>\hat\tau_\alpha)$ is formally defined.
 This classifier ensures that, for a given significance level $\alpha$, the Type I error is controlled below $\alpha$ with high probability $1-\delta$. Consequently, when $\hat\eta(q,M(q)) \geq \hat\tau_\alpha$, we reject the null hypothesis $H_0$ and assert that the model $M$ can answer the question $q$ certainly and correctly. Otherwise, the model will output an acknowledgment of uncertainty. 
 

\subsection{Type II Error Control}\label{sec:power}

The effectiveness of \approach not only hinges on Type I error control but also on ensuring sufficient statistical power to detect true positives. We then analyze the Type II error of the constructed classifier.

Denote $\eta(q',M(q'))=\Prob_{(q,M(q),y)\sim P}(y=1|q=q',M(q)=M(q'))$ to be the conditional probability that $M$ answers the question $q'$ certainly given any question $q'$ and the generated answer $M(q')$. Note that a question $q$ may have multiple correct answers and $a$ is just one realization. Therefore, $a$, and thus $y$, may still be random given $(q,M(q))$, implying $\eta(q,M(q))$ may take value in $(0,1)$.
Suppose there exists an increasing function $H$ such that $\|H\circ\hat \eta-\eta\|_\infty\le\epsilon_\eta$, where $H\circ\hat\eta(q,M(q))=H(\hat\eta(q,M(q)))$ is the composition of $H$ and $\hat\eta$. Note that $\hat k$, and thus $\hat f_\alpha$, are invariant if we replace $\hat\eta$ in Section \ref{sec:method} by $H\circ \hat \eta$. Therefore, without the loss of generality, we assume $H$ is the identity function. Denote $F(t)=\Prob_{(q,M(q))\sim P_0}(\hat\eta(q,M(q))\le t)$ to be the CDF of $\hat\eta(q,M(q))$ under $P_0$. For any classifier $f$, we set $\gR_0(f)=\Prob_{(q,M(q))\sim P_0}(f(q,M(q))=1)$ (resp. $\gR_1(f)=\Prob_{(q,M(q))\sim P_1}(f(q,M(q))=0)$) to be the Type I error (resp. Type II error). It follows from Theorem 1 in \cite{JMLR:v14:tong13a} that the Bayes optimal classifier $f_\alpha^*$
\[f_\alpha^*\in\argmin_{f:\mathcal{Q}\times\mathcal A\rightarrow\{0,1\}} \gR_1(f)\quad{\rm s.t.}\quad\gR_0(f)\le\alpha\]
has the form $f^*_\alpha(q,M(q))=\sI(\eta(q,M(q))>\tau_\alpha)$ for some $\tau_\alpha\in[0,1]$. 

Let $p_y=\Prob_{(q,M(q),y)\sim P}(y=1)$ denote the marginal probability that model $M$ is certain. We define 
\[\xi_\alpha=\frac{\tau_\alpha(1-p_y)}{(1-\tau_\alpha)p_y},\quad\alpha'=\alpha-c\sqrt{\frac{\alpha}{n_0}\log\frac{1}{\delta}},\quad\epsilon_\tau=\tau_{\alpha'}-\tau_{\alpha}+\epsilon_\eta,\]
for some constant $c>0$. If we denote $G_\alpha(\epsilon)=\Prob_{(q,M(q))\sim P_0}(|\eta(q,M(q))-\tau_\alpha|\le\epsilon)$ to be the probability measure around the classification boundary of $f^*_\alpha$, then the Type II error of the proposed algorithm can be controlled as follows.

\begin{Theorem}\label{thm:power}

    If $\hat\eta(q,M(q))$ is a continuous random variable with $(q,M(q))\sim P_0$ and $\tau_\alpha+\epsilon_\tau+\epsilon_\eta<1$, then with probability at least $1-2\delta$, we have
    \[\gR_1(\hat f_\alpha)-\gR_1(f^*_{\alpha})\lesssim \xi_\alpha\sqrt{\frac{\alpha}{n_0}\log\frac{1}{\delta}}+\frac{(1-p_y)(\epsilon_\tau+\epsilon_\eta)}{p_y(1-\tau_\alpha-\epsilon_\tau-\epsilon_\eta)^2}G_{\alpha}(\epsilon_\tau+\epsilon_\eta).\]
\end{Theorem}

\section{Extension of \approach to Covariate Shifts}
\label{sec:extension}
The threshold selection procedure developed in Section \ref{sec:method} relies on the assumption that the calibration
 dataset $\D_0=\{(q_i,M(q_i))\in\QQ\mid y_i=0\}$ follows the target distribution $P_0$ of uncertain question-generated answer pairs. However, labeled samples from the target distribution may not always be available in practice. Instead, people may use the labeled data that they believe to be similar to the target distribution, which necessitates methods to handle distribution shifts.
In this section, we study the case of covariate shift, where the distribution of the question-generated answer pairs in the calibration data differs from that in the target distribution, while the conditional distribution of $y$ given $(q,M(q))$ remains the same.

\subsection{Setup}
Suppose we observe $n$ samples $\D=\{(q_i,M(q_i),y_i):i\in[n]\}$ from the source distribution $\tilde P$. We assume $P_{y|q,M(q)}=\tilde P_{y|q,M(q)}$ but $P_{q,M(q)}\ne\tilde P_{q,M(q)}$. Following Section \ref{sec:method}, we split $\D$ into a certain subset $\D_1=\{(q_i,M(q_i)): y_i=1, i\in[n]\}=\{(q_i^{(1)},M(q_i^{(1)})):i\in[n_1]\}$ and an uncertain subset $\D_0=\{(q_i,M(q_i)): y_i=0,i\in[n]\}=\{(q_i^{(0)},M(q_i^{(0)})):i\in[n_0]\}$. We denote the distribution of $\D_0,\D_1$ to be $\tilde P_0,\tilde P_1$, respectively. We further denote the density ratio between the target distribution $P_0$ of uncertain question-generated answer pair and the source distribution $\tilde P_0$ to be $w(q,M(q))=\frac{dP_0}{d\tilde P_0}(q,M(q))$. In this section, we assume $w$ is known and satisfies $w(q,M(q))\le B$ for all $(q,M(q))\in\QQ\times\mathcal{A}$.

\subsection{Type I Error Control under Covariate Shift}
To extend the procedure in Section \ref{sec:method} to the covariate shift setting, we take an additional step to transform the samples in $\D_0$ from $\tilde P_0$ to $P_0$ distributed random variables by rejection sampling.

In the first step, we generate $n_0$ uniform random variables $U_1,\ldots,U_{n_0}\overset{i.i.d.}{\sim} {\rm Unif}[0,B]$ and select the indexes $\II=\{i\in[n_0]:U_i\le w(q_i^{(0)},M(q_i^{(0)}))\}$. If we collect all the samples in $\D_0$ with indexes in $\II$ to form $\tilde\D_0=\{(q_i^{(0)},M(q_i^{(0)})):i\in\II\}\overset{\triangle}{=}\{(\tilde q_i,M(\tilde q_i)):i\in[\tilde n_0]\}$. Then it can be shown that $\tilde\D_0\mid\II\overset{\rm i.i.d.}{\sim} P_0$.

In the second step, we apply the procedure introduced in Section \ref{sec:method} to the uncertain subset $\tilde \D_0$. Specifically, given the uncertain subset $\tilde \D_0$, we calculate the certainty scores $\tilde T_i=\hat\eta(\tilde q_i,M(\tilde q_i))$ and order them in increasing order to get $\tilde T_{(1)}\le\ldots\le \tilde T_{(\tilde n_0)}$, and set $\tilde T_{(\tilde n_0+1)}=+\infty$. Then we set the threshold $\hat\tau_\alpha$ to be $\tilde T_{(\hat k)}$, with $\hat k$ satisfies
\[\hat k=\min\{k\in[\tilde n_0+1]:\tilde v(k)\le\delta\},\quad \tilde v(k)=\sum_{j=k}^{\tilde n_0}{\tilde n_0\choose j}(1-\alpha)^j\alpha^{\tilde n_0-j},\quad \tilde v(\tilde n_0+1)=0.\]
Then we are able to control the Type I error as follows.

\begin{Theorem}\label{thm:type1_shift}
    With probability at least $1-\delta$, the constructed classifier $\hat f_\alpha(q,M(q))=\sI(\hat\eta(q,M(q))>\tilde T_{(\hat k)})$ has type I error below $\alpha$, i.e.
    \[\Prob_\D(\Prob_{(q,M(q))\sim P_0}(\hat f_\alpha(q,M(q))=1)\le\alpha)\ge1-\delta.\vspace{-5pt}\]
\end{Theorem}

\vspace{0pt}\section{Experiments}\label{sec:experiment}

In this section, we empirically investigate \approach in addressing the hallucination problem
of LLMs, focusing on the following questions: \highlight{\textbf{Q1:}} Can \approach improve the accuracy and lead to more factual LLMs? \highlight{\textbf{Q2:}} Can \approach effectively control the Type I error? \highlight{\textbf{Q3:}} Can \approach generalize well when covariate shifts exist?

\subsection{Experimental Setups}
\label{sec:setup}
\textbf{Datasets.} 
Following R-Tuning~\citep{zhang2023r}, we conduct our experiments on the knowledge-extensive QA tasks, which can then be categorized into two generation tasks. More details about the datasets can be referred to Appendix~\ref{sec:datasets}.

\begin{itemize}[left=1pt]
    \item \textit{Question-Answering: }Given a question, the model directly predicts its answer, which is one sentence with tokens no longer than 15. We include \textbf{ParaRel}~\citep{elazar2021measuring} and \textbf{HotpotQA}~\citep{yang2018hotpotqa}. For experiments considering distirbution shifts, we utilize \textbf{ParaRel-OOD} as the testing dataset.
    \item \textit{Multiple-Choice: }Given a question with several choices, the model chooses one option among A, B and C. We include \textbf{WiCE}~\citep{kamoi2023wice} and \textbf{FEVER}~\citep{thorne2018fever}.
\end{itemize}

\textbf{Certainty Score Functions.} 
We can either fit a prediction model to predict the certainty of a given question or use any off-the-shelf certainty estimation function to serve as the score function $\hat\eta$. Particularly, in this paper, we introduce three entropy-based certainty functions. Details about the score functions and equations are deferred to Appendix~\ref{sec:certain}.

\begin{itemize}[left=1pt]
    \item \textbf{Vanilla Entropy (VE)}: We query the model $M$ $k$ times and calculate the entropy across $k$ answers. 
    \begin{equation}
    VE(q, M(q)) = - \sum_{j=1}^k p(M(q)_j|q) \log p(M(q)_j|q),\ \hat\eta(q,M(q)) = -VE(q,M(q)).
    \end{equation}
    where $p(M(q)_j|q)$ is the frequency of a predicted answer $M(q)_j$ given a question $q$. 
    \item \textbf{Semantic Entropy (SE)}: \citet{semantic} measures uncertainty in natural language generation by accounting for the probability distribution over distinct meanings rather than individual token sequences.
    \item \textbf{Kernel Language Entropy (KLE)}: \citet{kle} quantifies uncertainty by using semantic similarity kernels between generated answers, allowing for a more nuanced estimation of uncertainty. Notably, this function does not apply to multiple-choice datasets and we only employ it on ParaRel and HotpotQA.
    
\end{itemize}

\textbf{Models.} In main experiments, we focus on distribution-free settings, where models that require fine-tuning will violate this assumption. We implement seven variants of \approach, utilizing three score functions, and compare their performances against baseline pretrained models. The baseline model, denoted as \textit{Pretrained}, involves evaluating the original pretrained checkpoints on the entire test set without any modifications. In contrast, \approach models are assessed solely on questions for which they can confidently provide answers. Among them, $\text{\approach-ve}_{k}$, $\text{\approach-se}_{k}$ and $\text{\approach-kle}_{k}$ means utilizing vanilla entropy, semantic entropy and kernal language entropy as score functions, respectively, where $k$ means we sample $k$ outputs for a given question.

To facilitate comparison with training-based methods, we randomly split our training dataset, allocating half for instruction-tuning and the remaining half to construct the calibration dataset. We choose semantic entropy as the score function and generate 15 outputs. This variant is denoted as \approach-t.
For comparative analysis, we include \textit{R-Tuning}~\citep{zhang2023r} as our primary baseline, evaluating it on the subset of questions that it is willing to answer. We also consider \textit{Finetune-All} and \textit{Finetune-Half}, which undergo instruction-tuning using the entire and half of the original training dataset, respectively, and are evaluated on the entire test set.

To assess the applicability of our framework on black-box APIs, we further implement \approach on GPT-4o Mini~\citep{gpt4omini} and compare the resulting performances.

\textbf{Metrics.} 
For models that could only output either the answer or an unknown expression, we evaluate the questions that our model is willing to answer. The accuracy is calculated as follows:
\begin{equation}
    \text{Acc} = \frac{\text{\# of correctly and willingly answered questions}}{\text{\# of willingly answered questions}}.
\end{equation}

Besides, we also include Type I error, or False Positive Rate (FPR), and Type II error, or False Negative Rate (FNR), as our evaluation metrics. Notably, our method enforces an upper bound of Type I error, that is, the probability of predicting an uncertain question as a certain one.

\textbf{Implementation.} 
We choose OpenLLaMA-3B, OpenLLaMA-7B, OpenLLaMA-13B~\citep{openlm2023openllama}, and LLaMA-7B, LLaMA-13B~\citep{touvron2023llama} as the base models in our experiments. The temperature is set to $0$ for evaluation and $0.7$ for calculating score functions. We follow \citet{zhang2023r} to use LMFlow~\citep{diao2023lmflow} to conduct instruction tuning, setting epoch to 1 and learning rate to $2e^{-5}$.
All the experiments are implemented on 4 Nvidia H100-80GB GPUs.

\subsection{Main Experimental Results}

\begin{table*}[tb!]
\centering
    \captionsetup{font={footnotesize}}
    \caption{The accuracy performance (\%) of \approach compared to Pretrained models on question-answering and multiple-choice datasets using a significance level of $\alpha = 0.05$. For brevity, \approach is abbreviated as \mini. The notation $\text{\mini-ve}_{15}$ denotes the use of a vanilla entropy score function with 15 generated outputs. }\vspace{-5pt}
    \label{tab:main_acc}
    \resizebox{\linewidth}{!}{
\begin{tabular}{c|c|c|ccc|ccc|c}
\toprule[1.0pt]
Dataset                   & Model                & Pretrained       & $\text{\mini-ve}_5$  & $\text{\mini-ve}_{10}$  & $\text{\mini-ve}_{15}$  & $\text{\mini-se}_5$  & $\text{\mini-se}_{10}$ & $\text{\mini-se}_{15}$ & $\text{\mini-kle}_{15}$ \\
\midrule[0.6pt]
\multirow{3}{*}{ParaRel}  & OpenLLaMA-3B   & \baseline{36.66} & 60.54  & 66.75 & \secondplace{67.28} & 60.10 & 62.50 & \thirdplace{67.26} & \firstplace{78.45}\\
                          & OpenLLaMA-7B   & \baseline{40.38} & 74.92 & \secondplace{79.87} & \firstplace{80.29} & 65.53 & 71.40 & 65.23 & \thirdplace{76.83} \\
                          & OpenLLaMA-13B  & \baseline{42.21}  & \thirdplace{77.37} & 77.31 & \secondplace{79.41} & 73.49 & 68.89 & 73.09 & \firstplace{83.84}\\
\midrule[0.6pt]
\multirow{3}{*}{HotpotQA} & OpenLLaMA-3B   & \baseline{25.72}  & 50.81 & \secondplace{55.19} & \thirdplace{53.75} & 45.37 & 52.55 & 52.66 & \firstplace{55.35} \\
                          & OpenLLaMA-7B   & \baseline{28.63} & 56.06 & \thirdplace{59.69} & \firstplace{60.67} & 51.48 & 53.75 & 56.56 & \secondplace{60.66}\\
                          &  LLaMA-13B  & \baseline{30.83}  & 51.49 & 54.41 & 49.74 & \thirdplace{55.41} & \secondplace{57.18} & \firstplace{60.69} & 54.49\\
\midrule[0.6pt]
\multirow{3}{*}{WiCE}     & OpenLLaMA-3B   & \baseline{64.72} & 67.65 & \secondplace{75.00} & \thirdplace{68.18} & 64.71 & \firstplace{85.71} & 66.67 & -- \\
                          & OpenLLaMA-7B   & \baseline{72.96} & 50.00 & 55.88 & 47.37 & \secondplace{90.00} & \firstplace{100.0} & \secondplace{90.00} & --\\
                          &  LLaMA-13B  & \baseline{56.89} & 63.33 & 45.45 & 44.44 & \firstplace{100.0} & \thirdplace{82.35} & \secondplace{90.00} & --\\
\midrule[0.6pt]
\multirow{3}{*}{FEVER}    & OpenLLaMA-3B   & \baseline{39.74} & 60.24 & 62.50 & 41.72  & \secondplace{82.40} & \thirdplace{79.23} & \firstplace{83.90}  & --\\
                          &     LLaMA-7B & \baseline{35.99} & \thirdplace{43.92} & \secondplace{50.94} & \firstplace{51.38} & 28.69 & 33.12 & 33.27 & --\\
                          &  LLaMA-13B & \baseline{32.15} & 38.74 & 42.48 & 46.07 & \thirdplace{49.92} & \firstplace{54.17} & \secondplace{52.23} & --\\
\bottomrule[1.0pt]
\end{tabular}}
\end{table*}

We first conduct in-distribution experiments on question-answering and multiple choice datasets ParaRel, HotpotQA, WiCE and FEVER. 

\textbf{Main Performance.} The accuracy results are presented in Table~\ref{tab:main_acc}, where the significance level $\alpha$ for \approach is set to 0.05. Additional experimental results for other significance levels (e.g., $\alpha = 0.10$) are provided in Appendix~\ref{sec:morealpha}. Analysis of the results reveals that \approach significantly outperforms pretrained models by a substantial margin in terms of accuracy on the questions it is willing to answer, compared to baselines that respond to all questions indiscriminately. Notably, \approach can yield an over 40\% accuracy improvement on ParaRel, WiCE and FEVER. These results demonstrate that \approach is more reliable when making predictions and is capable of refusing unknown answers.

\begin{table*}[tb!]
\centering
    \caption{The Type I error of Fact-Test on question-answering and multiple-choice datasets. The significance level $\alpha=0.05$.}\vspace{-5pt}
    \label{tab:fnr_table}
    \resizebox{0.9\linewidth}{!}{
\begin{tabular}{c|c|ccc|ccc|c}
\toprule[1.0pt]
Dataset                   & Model                &  $\text{\mini-ve}_5$  & $\text{\mini-ve}_{10}$  & $\text{\mini-ve}_{15}$  & $\text{\mini-se}_5$  & $\text{\mini-se}_{10}$ & $\text{\mini-se}_{15}$ & $\text{\mini-kle}_{15}$  \\
\midrule[0.6pt]
\multirow{3}{*}{ParaRel}  & OpenLLaMA-3B   & 0.0455  & 0.0467 & 0.0513 & 0.0479 & 0.0520 & 0.0486 & 0.0342 \\
                          & OpenLLaMA-7B   & 0.0225 & 0.0093 & 0.0145 & 0.0393 & 0.0394 & 0.0435 & 0.0400\\
                          & OpenLLaMA-13B  & 0.0192  & 0.0087 & 0.0302 & 0.0341 & 0.0477 & 0.0337 & 0.0331\\
\midrule[0.6pt]
\multirow{3}{*}{HotpotQA} & OpenLLaMA-3B   & 0.0276  & 0.0250 & 0.0268 & 0.0321 & 0.0287 & 0.0291 & 0.0311 \\
                          & OpenLLaMA-7B   & 0.0295  & 0.0298 & 0.0289 & 0.0310 & 0.0325 & 0.0291 & 0.0293\\
                          &  LLaMA-13B  & 0.0222 & 0.0232 & 0.0378 & 0.0273 & 0.0296 & 0.0244 & 0.0316\\
\midrule[0.6pt]
\multirow{3}{*}{WiCE}     & OpenLLaMA-3B   & 0.0325 & 0.0089 & 0.0207 & 0.0175 & 0.0029 & 0.0118 & --\\
                          & OpenLLaMA-7B   & 0.0694 & 0.0579 & 0.0617 & 0.0077 & 0.0 & 0.0039  & --\\
                          &  LLaMA-13B  & 0.0266 & 0.0290 & 0.0363 & 0.0 & 0.0072 & 0.0024  & --\\
\midrule[0.6pt]
\multirow{3}{*}{FEVER}    & OpenLLaMA-3B   & 0.0164 & 0.0005 & 0.0217  & 0.0570 & 0.0471 & 0.0496 & -- \\
                          &     LLaMA-7B & 0.0598  & 0.0081 & 0.0329 & 0.0392 & 0.0495 & 0.0495 & --\\
                          & LLaMA-13B & 0.0172 & 0.0383 & 0.0293  & 0.0459 & 0.0518 & 0.0552 & -- \\
\bottomrule[1.0pt]
\end{tabular}}
\end{table*}

\begin{figure}[tb!]
\centering
\subfigcapskip=-5pt
  \subfigure[ParaRel-3B]{
    {\includegraphics[width=0.3\linewidth]{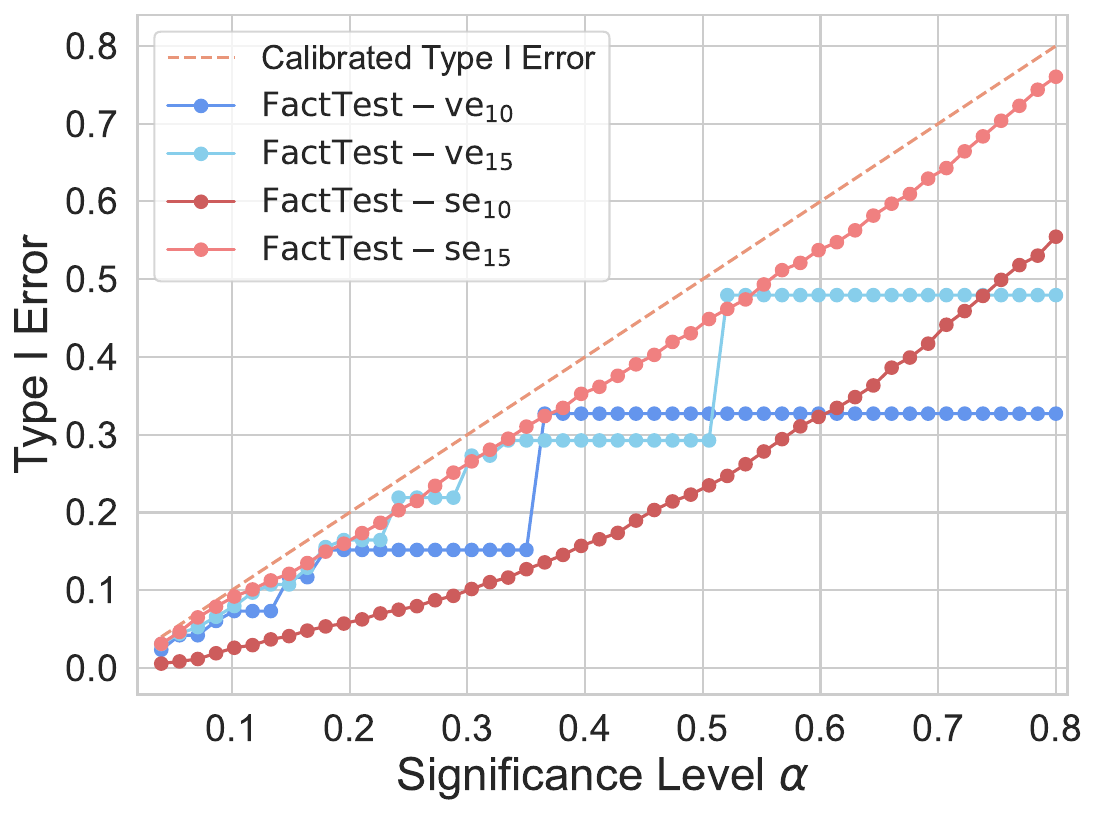}}}\hspace{5pt}
  \subfigure[WiCE-3B]{
    {\includegraphics[width=0.3\linewidth]{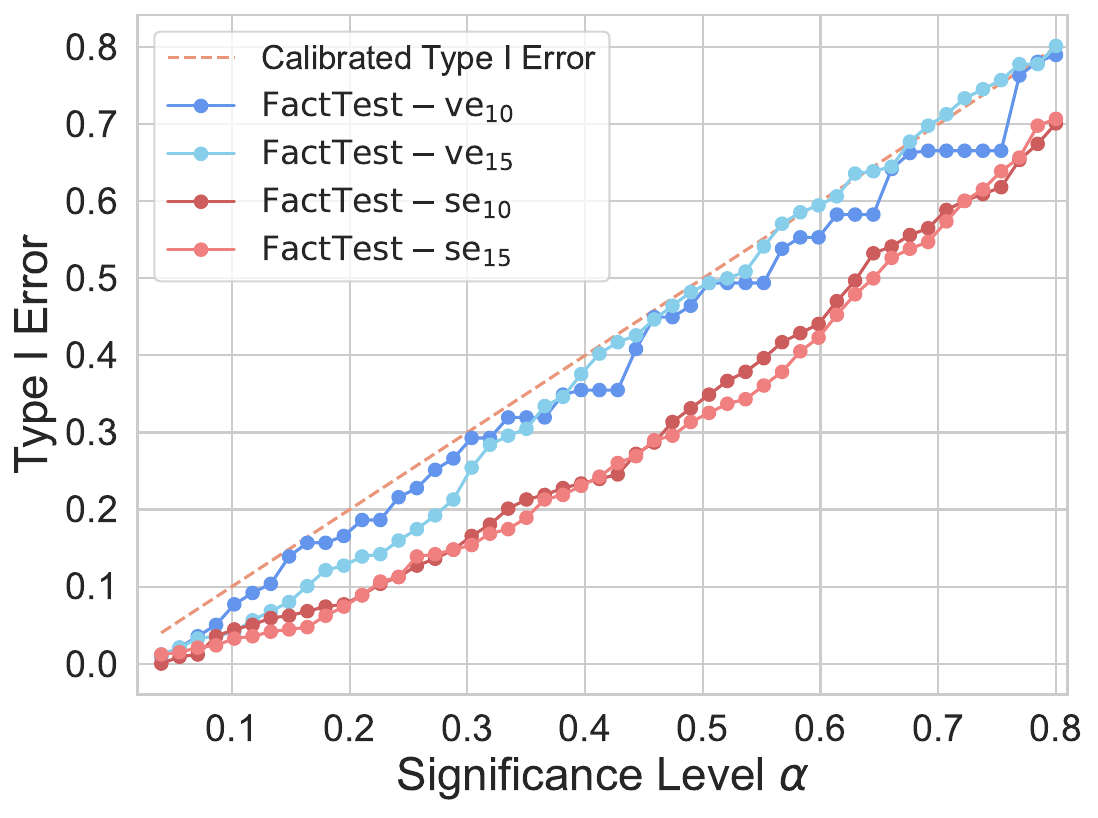}}}\hspace{5pt}
  \subfigure[FEVER-3B]{
    {\includegraphics[width=0.3\linewidth]{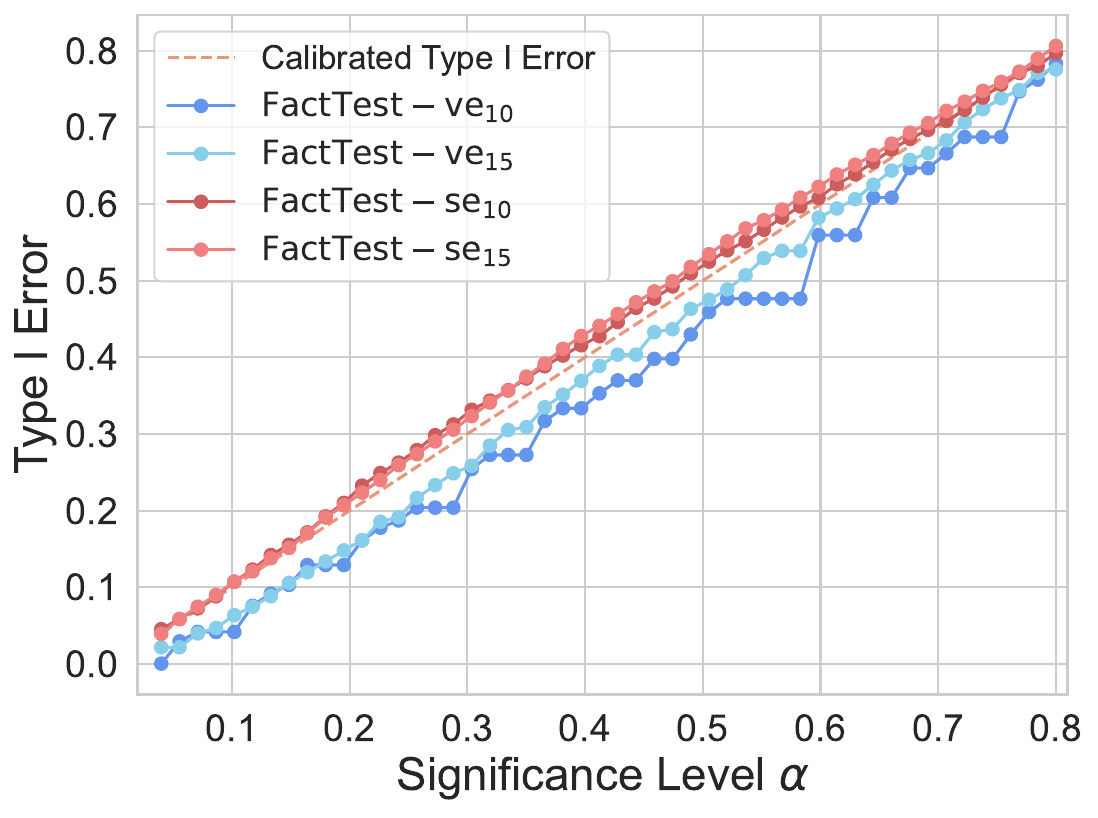}}} 
    \\ \vspace{-7pt}
 \caption{\approach can control the Type I error given a significance level $\alpha$. The caption of each sub-figure consists of the dataset name and the model size. }
  \vspace{-20pt}
\label{fig:main_fnr}
\end{figure}

\textbf{Type I Error.} Table~\ref{tab:fnr_table} represents the Type I error, or FPR, of \approach when $\alpha$ is set to 0.05. Figure~\ref{fig:main_fnr} depicts the FPR-$\alpha$ curve. For a given significance level $\alpha$, we enforce an upper bound on the FPR at $\alpha$ with a high probability guarantee. Analysis of these figures confirms that our method reliably controls the Type I error, thereby validating the theoretical results presented in Section~\ref{section:calibration}. Due to space constraints, additional error control results including more Type I Error results and Type II Error simulation for \approach are available in Appendix~\ref{sec:error}.

\begin{figure}[tb!]
\centering
\subfigcapskip=-5pt
  \subfigure[ParaRel]{
    {\includegraphics[width=0.9\linewidth]{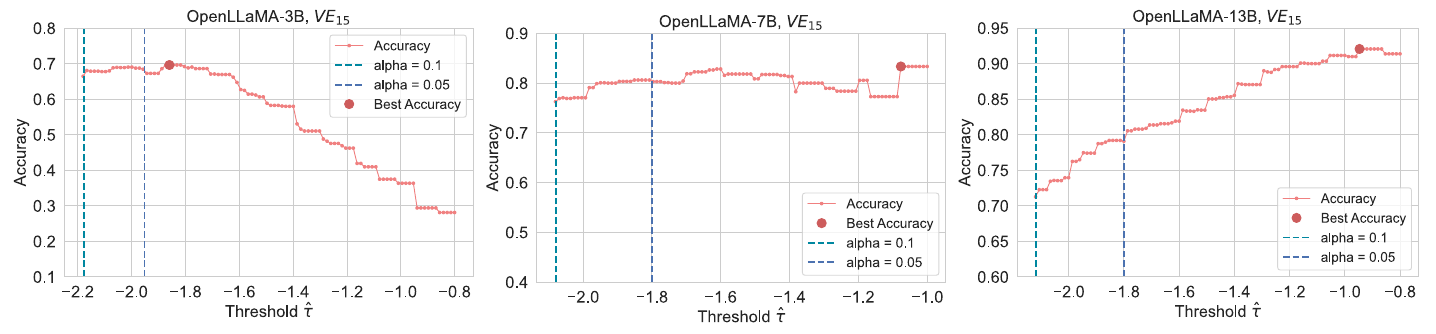}}}\hspace{5pt}
  \subfigure[WiCE]{
    {\includegraphics[width=0.9\linewidth]{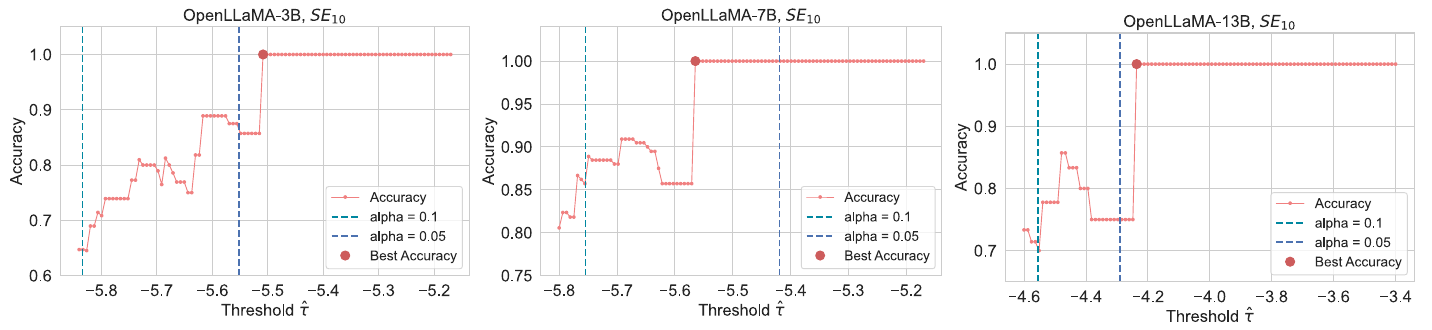}}} 
    \\ \vspace{-7pt}
 \caption{The Accuracy-Threshold curve. The title of each sub-figure consists of the dataset name, the model size and the certainty function. }
\label{fig:main_max}
\end{figure}

\textbf{Maximizing Accuracy.} Given a significance level $\alpha$, we can determine the threshold $\hat{\tau}_\alpha$ that minimizes Type II error while ensuring that the Type I error remains within the specified upper bound. For $\hat{\tau} > \hat{\tau}_\alpha$, the Type I error decreases monotonically, whereas the Type II error increases monotonically. Figure~\ref{fig:main_max} presents the accuracy-$\hat{\tau}$ curve, where $\hat{\tau}$ begins at $\hat{\tau}_{0.1}$. This curve can be utilized to maximize accuracy, which does not follow a monotonic trend as the threshold $\hat{\tau}$ increases, while ensuring that the Type I error is controlled below 0.10.

\subsection{Comparing with Finetuned Models}\label{sec:rtuning}

\begin{figure}[tb!]
\centering
\includegraphics[width=0.9\linewidth]{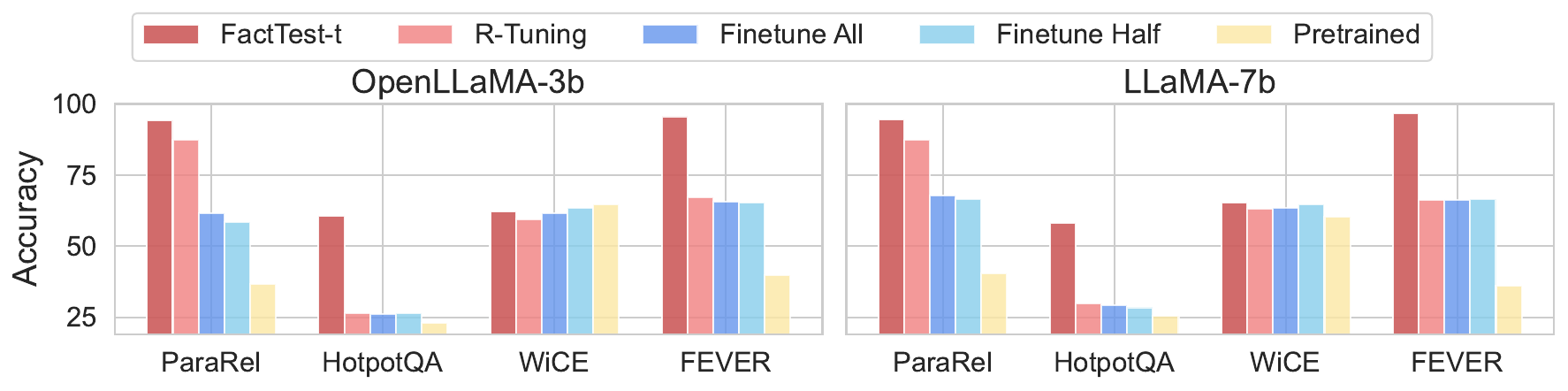}
\vspace{-5pt}
 \caption{The Accuracy performance (\%) of \approach trained on half of the data, comparing with training-based baselines. Both R-Tuning and Finetune All utilize all training data for finetuning, while Finetune Half uses the same half of the finetuning data as \approach.}
  \vspace{-10pt}
\label{fig:rtuning}
\end{figure}

Figure~\ref{fig:rtuning} illustrates the accuracy performance of \approach-t compared to the baseline methods R-Tuning, Finetune-All, and Finetune-Half. We randomly divide $\D$ into two equal parts: $\mathcal{D}_{I}$ for instruction-tuning and $\mathcal{D}_{C}$ for constructing the calibration dataset. The pretrained model is finetuned on $\mathcal{D}_{I}$ to obtain Finetune-Half, while Finetune-All is obtained by training on the entire dataset $\mathcal{D}$. For R-Tuning, we also utilize the entire dataset to finetune the model. It is evident that \approach-t consistently outperforms R-Tuning by a large margin, while utilizing only half of the available training data, thereby reducing training costs by 50\%. Notably, \approach-t yields 34\% and 28\% accuracy improvement over R-Tuning on HotpotQA and FEVER, respectively. Despite the reduced size of the calibration dataset, \approach-t maintains effective control over Type I error, with further details provided in Appendix~\ref{sec:error}. Answer rate analysis of \approach-t and baselines is also provided in Appendix~\ref{sec:answer}.

\subsection{Extension to Covariate Shifts}
\begin{figure}[tb!]
\centering
\includegraphics[width=0.9\linewidth]{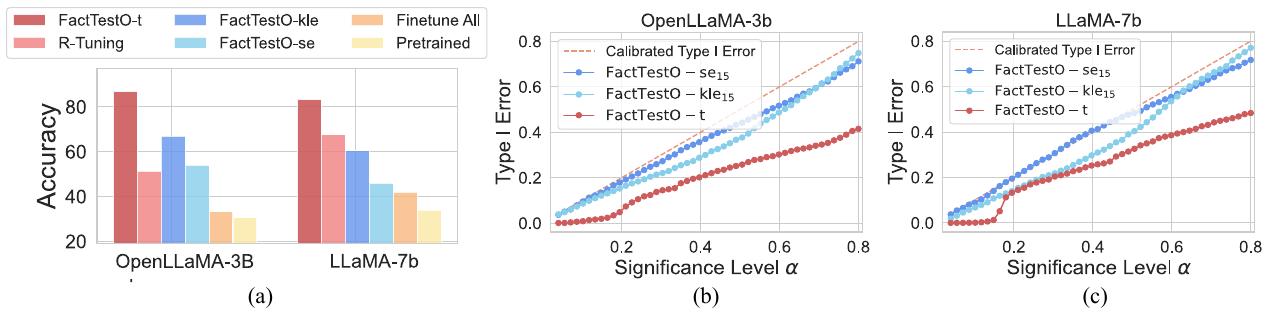}
 \caption{(a) The Accuracy performance (\%) of \approach on ParaRel-OOD testing dataset. (b)(c) \textsc{FactTestO} maintains its ability to control Type I error given a significance level $\alpha$ when distribution shifts exist.}

\label{fig:ood}
\end{figure}

In this subsection, we evaluate the extension of our framework, denoted as \textsc{FactTestO} (\approach for \textsc{O}ut-of-distribution domains), on the dataset containing distribution shifts. 

\textbf{Setup.} We utilize ParaRel as the training dataset, consistent with the aforementioned experiments. We randomly split ParaRel-OOD into a validation dataset comprising 1,000 samples and a testing dataset containing 11k samples. As outlined in Section~\ref{sec:extension}, it is necessary to calculate the density ratio between the target distribution and the source distribution. To achieve this, we employ the training data from the source domain and the validation data from the target domain to train a classifier for approximating density ratios. Subsequently, we select $B$ as the $\gamma$ upper quantile of density ratios to filter out anomalous values. We set the default value of $\gamma$ as 90\%.

\textbf{Experimental Results.} 
Figure~\ref{fig:ood} depicts the accuracy performance of \textsc{FactTestO} on the ParaRel-OOD testing dataset, alongside the Type I error-$\alpha$ curve. The results demonstrate that \textsc{FactTestO}-t significantly outperforms baseline methods by a large margin. Notably, when utilizing OpenLLaMA-3B as the pretrained model, both \approach-se and \approach-kle outperform training-based methods without requiring fine-tuning. Additionally, the figures confirm that the \textsc{FactTestO} extension effectively enforces the upper bound on the Type I error, thereby maintaining its efficacy in out-of-distribution scenarios.

\begin{table*}[tb!]
\centering
    \caption{The accuracy performance (\%) of \approach applied to GPT-4o-mini. The significance level is chosen as 5\%. The number in parentheses is \textcolor{myblue}{Type I error}. GPT + OpenLLaMA-7B means utilizing OpenLLaMA-7B to calculate certainty scores for GPT-4o mini.}\vspace{-5pt}
    \label{tab:gpt}
    \resizebox{0.7\linewidth}{!}{
\begin{tabular}{c|c|ccc}
\toprule[1.0pt]
Dataset                   & Model                &  $\text{\mini-se}_{10}$  & $\text{\mini-se}_{15}$  & $\text{\mini-kle}_{15}$ \\
\midrule[0.6pt]
\multirow{3}{*}{ParaRel}& GPT-4o mini & \multicolumn{3}{c}{\baseline{52.83}} \\
& GPT + OpenLLaMA-7B & 77.78 (\textcolor{myblue}{0.03}) & 77.31 (\textcolor{myblue}{0.03})& 83.88 (\textcolor{myblue}{0.05})\\
& GPT + OpenLLaMA-13B & 76.91 (\textcolor{myblue}{0.04}) & 77.67 (\textcolor{myblue}{0.05}) & 85.84 (\textcolor{myblue}{0.04})\\
\midrule[0.6pt]
\multirow{3}{*}{WiCE}& GPT-4o mini & \multicolumn{3}{c}{\baseline{75.67}} \\
& GPT + OpenLLaMA-7B & 81.82 (\textcolor{myblue}{0.02}) & 76.67 (\textcolor{myblue}{0.03}) & --\\
& GPT + OpenLLaMA-13B & 86.95 (\textcolor{myblue}{0.01}) & 81.77 (\textcolor{myblue}{0.02}) & --\\
\bottomrule[1.0pt]
\end{tabular}}
\end{table*}

\subsection{Extension to Black-box APIs}\vspace{-5pt}
We further evaluate our framework on black-box models, such as GPT-4o Mini~\citep{gpt4omini}, to broaden the applicability of our framework. While certainty functions like SE and KLE require token probabilities, which are unavailable for black-box APIs, we utilize open-source models to calculate the certainty scores on calibration datasets constructed by black-box models. Table~\ref{tab:gpt} illustrates the performance of \approach on GPT-4o Mini. The results demonstrate that the certainty scores derived from open-source models are effective for black-box APIs, achieving a 33\% accuracy improvement on ParaRel and an 11\% improvement on WiCE, while maintaining control over Type I error. These findings illustrate that our framework provides a practical and effective solution for detecting hallucinations in closed-box models.

\section{Related Work}\label{sec:related}

\textbf{Factuality of LLMs.} The factuality of LLMs is a major problem and of significant research interest since these models are likely to hallucinate unfaithful and nonfactual facts, which severely impacts the reliability and trustworthiness of LLMs~\citep{ji2023survey, maynez-etal-2020-faithfulness,li-etal-2023-defining}.

Recently, a variety of works have been done towards hallucination detection, mitigation and evaluation~\citep{huang2023survey, wang2023survey}. Our work relates more to hallucination detection, which is imperative for assuring the reliability of the generated content. 
Retrieving-based approaches
\citep{kadavath2022language} proposes self-evaluation by asking models to first propose answers, and then evaluate their previous prediction. \citet{azaria2023internal} trains a classifier based on hidden layer activations, which is a white-box approach that requires internal states of the LLMs.
\citet{lee2023factuality} uses factual and nonfactual prompts and creates a benchmark for measuring the factuality of generations. 
\citet{manakul2023selfcheckgpt} introduces SelfCheckGPT to fact-check the responses of LLMs in a zero-resource fashion. \citet{zhang2023r} instructs LLMs to refuse unknown questions by refusal-aware instruction tuning. \citet{xu2024sayself} generate self-reflective rationales and teach LLMs to express fine-grained confidence estimates. However, none of these works have provided theoretical guarantees.



\textbf{Uncertainty Quantification of LLMs.}
Our work relates to a line of work on uncertainty quantification (UQ) for LLMs, as we employ certainty functions to assess models' ability to reliably give an answer. Predictive entropy that measures the entropy of the model's predicted token distribution has been used as a simple baseline for UQ in LLMs~\citep{braverman2019calibrationentropyratesmemory}. \citet{semantic} introduced Semantic Entropy, which incorporates linguistic invariances to measure uncertainty. Most recently, \citet{kle} introduced Kernel Language Entropy (KLE), which defines positive semidefinite unit trace kernels and quantifies uncertainty using the von Neumann entropy. \citet{lin2024generating} proposed a simple supervised approach for uncertainty estimation in black-box LLMs using labeled datasets. \citet{duan2024shifting} identifies that existing methods treat all tokens equally when estimating uncertainty and proposes jointly Shifting Attention to more Relevant (SAR) components.

These works are complementary to ours, as our contribution is a meta-algorithm that works with any uncertainty quantification method to serve as the certainty score functions
and assess the factuality. Future developments in this line of work can greatly improve the performance of our framework.


\textbf{Neyman-Pearson Classification.} Instead of minimizing a weighted combination of Type I and II errors, as studied in standard classification and cost-sensitive learning, the Neyman-Pearson classification paradigm focuses on the setting where we prioritize the Type I error control and aim to achieve minimum Type II error while keeping the Type I error below a user-specified level $\alpha$. To achieve this goal, \cite{rigollet2011neyman,scott2005neyman} propose to use the empirical risk minimization strategy and \cite{JMLR:v14:tong13a} utilizes the plug-in approaches to construct the NP classifier. In a more related work \cite{tong2018neyman}, the authors propose an umbrella algorithm that achieves Type I error control for any pretrained classifiers. However, this work \citep{tong2018neyman} does not provide any Type II error guarantee for the proposed algorithm.

Our work takes an initial step to use the NP classification idea to do the factuality testing for LLMs. And the Type II error analysis for our method can be directly applied to the standard NP umbrella algorithm, which is of independent interest. Moreover, we extend the NP classification framework to the case with covariate shifts, enabling us to address more practical and real-world problems.

\section{Conclusion: Summary and Limitations}\label{sec:conclusion}
\textbf{Summary. }In this paper, we introduced \approach, a novel framework for factuality testing in Large Language Models (LLMs) that leverages the principles of Neyman-Pearson (NP) classification to provide finite-sample and distribution-free guarantees. By formulating factuality testing as a hypothesis testing problem, \approach effectively enforces an upper bound on Type I errors. We prove that our framework ensures strong power control under mild conditions and can be extended to maintain its effectiveness in the presence of covariate shifts. These theoretical analyses can be seamlessly integrated with the standard NP umbrella algorithm, not limited to our framework. Our approach is distribution-free and works for any number of human-annotated samples. It  applies to any LLM including closed-box models. Empirical evaluations have demonstrated that \approach consistently outperforms both pretrained and fine-tuned baselines by a large margin. Besides, \approach can be extended to maintain superior performance under distribution shifts, ensuring its robustness and reliability in real-world scenarios. Additionally, our framework effectively enhances the reliability of black-box APIs, highlighting its practical applicability.

\textbf{Limitation. }One limitation of our work is the current implementation of only three entropy-based certainty functions, which may not be optimal choices in every task. Exploring additional certainty estimation methods could further enhance the framework's performance. Furthermore, our framework constructs the certainty predictor in an offline manner. Future work could extend \approach to support online testing, thereby enabling real-time factuality assessments.

\newpage

\bibliography{iclr2025_conference}

\begin{thebibliography}{38}
\providecommand{\natexlab}[1]{#1}
\providecommand{\url}[1]{\texttt{#1}}
\expandafter\ifx\csname urlstyle\endcsname\relax
  \providecommand{\doi}[1]{doi: #1}\else
  \providecommand{\doi}{doi: \begingroup \urlstyle{rm}\Url}\fi

\bibitem[Azaria \& Mitchell(2023)Azaria and Mitchell]{azaria2023internal}
Amos Azaria and Tom Mitchell.
\newblock The internal state of an llm knows when it's lying, 2023.

\bibitem[Braverman et~al.(2019)Braverman, Chen, Kakade, Narasimhan, Zhang, and
  Zhang]{braverman2019calibrationentropyratesmemory}
Mark Braverman, Xinyi Chen, Sham~M. Kakade, Karthik Narasimhan, Cyril Zhang,
  and Yi~Zhang.
\newblock Calibration, entropy rates, and memory in language models, 2019.
\newblock URL \url{https://arxiv.org/abs/1906.05664}.

\bibitem[Chen et~al.(2024)Chen, Kim, Sriram, Durrett, and
  Choi]{chen2024complexclaimverificationevidence}
Jifan Chen, Grace Kim, Aniruddh Sriram, Greg Durrett, and Eunsol Choi.
\newblock Complex claim verification with evidence retrieved in the wild, 2024.
\newblock URL \url{https://arxiv.org/abs/2305.11859}.

\bibitem[Diao et~al.(2023)Diao, Pan, Dong, Shum, Zhang, Xiong, and
  Zhang]{diao2023lmflow}
Shizhe Diao, Rui Pan, Hanze Dong, Ka~Shun Shum, Jipeng Zhang, Wei Xiong, and
  Tong Zhang.
\newblock Lmflow: An extensible toolkit for finetuning and inference of large
  foundation models, 2023.

\bibitem[Duan et~al.(2024)Duan, Cheng, Wang, Zavalny, Wang, Xu, Kailkhura, and
  Xu]{duan2024shifting}
Jinhao Duan, Hao Cheng, Shiqi Wang, Alex Zavalny, Chenan Wang, Renjing Xu,
  Bhavya Kailkhura, and Kaidi Xu.
\newblock Shifting attention to relevance: Towards the predictive uncertainty
  quantification of free-form large language models.
\newblock In \emph{Proceedings of the 62nd Annual Meeting of the Association
  for Computational Linguistics (Volume 1: Long Papers)}, pp.\  5050--5063,
  2024.

\bibitem[Elazar et~al.(2021)Elazar, Kassner, Ravfogel, Ravichander, Hovy,
  Schütze, and Goldberg]{elazar2021measuring}
Yanai Elazar, Nora Kassner, Shauli Ravfogel, Abhilasha Ravichander, Eduard
  Hovy, Hinrich Schütze, and Yoav Goldberg.
\newblock Measuring and improving consistency in pretrained language models,
  2021.

\bibitem[Geng \& Liu(2023)Geng and Liu]{openlm2023openllama}
Xinyang Geng and Hao Liu.
\newblock Openllama: An open reproduction of llama, May 2023.
\newblock URL \url{https://github.com/openlm-research/open_llama}.

\bibitem[Gou et~al.(2024)Gou, Shao, Gong, Shen, Yang, Duan, and
  Chen]{gou2024criticlargelanguagemodels}
Zhibin Gou, Zhihong Shao, Yeyun Gong, Yelong Shen, Yujiu Yang, Nan Duan, and
  Weizhu Chen.
\newblock Critic: Large language models can self-correct with tool-interactive
  critiquing, 2024.
\newblock URL \url{https://arxiv.org/abs/2305.11738}.

\bibitem[Huang et~al.(2023)Huang, Yu, Ma, Zhong, Feng, Wang, Chen, Peng, Feng,
  Qin, and Liu]{huang2023survey}
Lei Huang, Weijiang Yu, Weitao Ma, Weihong Zhong, Zhangyin Feng, Haotian Wang,
  Qianglong Chen, Weihua Peng, Xiaocheng Feng, Bing Qin, and Ting Liu.
\newblock A survey on hallucination in large language models: Principles,
  taxonomy, challenges, and open questions, 2023.

\bibitem[Ji et~al.(2023)Ji, Lee, Frieske, Yu, Su, Xu, Ishii, Bang, Madotto, and
  Fung]{ji2023survey}
Ziwei Ji, Nayeon Lee, Rita Frieske, Tiezheng Yu, Dan Su, Yan Xu, Etsuko Ishii,
  Ye~Jin Bang, Andrea Madotto, and Pascale Fung.
\newblock Survey of hallucination in natural language generation.
\newblock \emph{ACM Computing Surveys}, 55\penalty0 (12):\penalty0 1--38, 2023.

\bibitem[Kadavath et~al.(2022)Kadavath, Conerly, Askell, Henighan, Drain,
  Perez, Schiefer, Hatfield-Dodds, DasSarma, Tran-Johnson, Johnston, El-Showk,
  Jones, Elhage, Hume, Chen, Bai, Bowman, Fort, Ganguli, Hernandez, Jacobson,
  Kernion, Kravec, Lovitt, Ndousse, Olsson, Ringer, Amodei, Brown, Clark,
  Joseph, Mann, McCandlish, Olah, and Kaplan]{kadavath2022language}
Saurav Kadavath, Tom Conerly, Amanda Askell, Tom Henighan, Dawn Drain, Ethan
  Perez, Nicholas Schiefer, Zac Hatfield-Dodds, Nova DasSarma, Eli
  Tran-Johnson, Scott Johnston, Sheer El-Showk, Andy Jones, Nelson Elhage,
  Tristan Hume, Anna Chen, Yuntao Bai, Sam Bowman, Stanislav Fort, Deep
  Ganguli, Danny Hernandez, Josh Jacobson, Jackson Kernion, Shauna Kravec,
  Liane Lovitt, Kamal Ndousse, Catherine Olsson, Sam Ringer, Dario Amodei, Tom
  Brown, Jack Clark, Nicholas Joseph, Ben Mann, Sam McCandlish, Chris Olah, and
  Jared Kaplan.
\newblock Language models (mostly) know what they know, 2022.

\bibitem[Kamoi et~al.(2023)Kamoi, Goyal, Rodriguez, and Durrett]{kamoi2023wice}
Ryo Kamoi, Tanya Goyal, Juan~Diego Rodriguez, and Greg Durrett.
\newblock Wice: Real-world entailment for claims in wikipedia, 2023.

\bibitem[Kuhn et~al.(2023)Kuhn, Gal, and Farquhar]{semantic}
Lorenz Kuhn, Yarin Gal, and Sebastian Farquhar.
\newblock Semantic uncertainty: Linguistic invariances for uncertainty
  estimation in natural language generation, 2023.
\newblock URL \url{https://arxiv.org/abs/2302.09664}.

\bibitem[Kumar et~al.(2023)Kumar, Lu, Gupta, Palepu, Bellamy, Raskar, and
  Beam]{kumar2023conformal}
Bhawesh Kumar, Charlie Lu, Gauri Gupta, Anil Palepu, David Bellamy, Ramesh
  Raskar, and Andrew Beam.
\newblock Conformal prediction with large language models for multi-choice
  question answering, 2023.
\newblock URL \url{https://arxiv.org/abs/2305.18404}.

\bibitem[Lee et~al.(2023)Lee, Ping, Xu, Patwary, Fung, Shoeybi, and
  Catanzaro]{lee2023factuality}
Nayeon Lee, Wei Ping, Peng Xu, Mostofa Patwary, Pascale Fung, Mohammad Shoeybi,
  and Bryan Catanzaro.
\newblock Factuality enhanced language models for open-ended text generation,
  2023.

\bibitem[Li et~al.(2023)Li, Han, Yu, Edwards, Li, Wang, Fung, Yu, Tetreault,
  Hovy, and Ji]{li-etal-2023-defining}
Sha Li, Chi Han, Pengfei Yu, Carl Edwards, Manling Li, Xingyao Wang, Yi~Fung,
  Charles Yu, Joel Tetreault, Eduard Hovy, and Heng Ji.
\newblock Defining a new {NLP} playground.
\newblock In Houda Bouamor, Juan Pino, and Kalika Bali (eds.), \emph{Findings
  of the Association for Computational Linguistics: EMNLP 2023}, pp.\
  11932--11951, Singapore, December 2023. Association for Computational
  Linguistics.
\newblock \doi{10.18653/v1/2023.findings-emnlp.799}.
\newblock URL \url{https://aclanthology.org/2023.findings-emnlp.799}.

\bibitem[Lin et~al.(2024)Lin, Trivedi, and Sun]{lin2024generating}
Zhen Lin, Shubhendu Trivedi, and Jimeng Sun.
\newblock Generating with confidence: Uncertainty quantification for black-box
  large language models, 2024.
\newblock URL \url{https://arxiv.org/abs/2305.19187}.

\bibitem[Manakul et~al.(2023)Manakul, Liusie, and
  Gales]{manakul2023selfcheckgpt}
Potsawee Manakul, Adian Liusie, and Mark~JF Gales.
\newblock Selfcheckgpt: Zero-resource black-box hallucination detection for
  generative large language models.
\newblock \emph{arXiv preprint arXiv:2303.08896}, 2023.

\bibitem[Maynez et~al.(2020{\natexlab{a}})Maynez, Narayan, Bohnet, and
  McDonald]{maynez-etal-2020-faithfulness}
Joshua Maynez, Shashi Narayan, Bernd Bohnet, and Ryan McDonald.
\newblock On faithfulness and factuality in abstractive summarization.
\newblock In Dan Jurafsky, Joyce Chai, Natalie Schluter, and Joel Tetreault
  (eds.), \emph{Proceedings of the 58th Annual Meeting of the Association for
  Computational Linguistics}, pp.\  1906--1919, Online, July
  2020{\natexlab{a}}. Association for Computational Linguistics.
\newblock \doi{10.18653/v1/2020.acl-main.173}.
\newblock URL \url{https://aclanthology.org/2020.acl-main.173}.

\bibitem[Maynez et~al.(2020{\natexlab{b}})Maynez, Narayan, Bohnet, and
  McDonald]{maynez2020faithfulness}
Joshua Maynez, Shashi Narayan, Bernd Bohnet, and Ryan McDonald.
\newblock On faithfulness and factuality in abstractive summarization.
\newblock \emph{arXiv preprint arXiv:2005.00661}, 2020{\natexlab{b}}.

\bibitem[Nikitin et~al.(2024)Nikitin, Kossen, Gal, and Marttinen]{kle}
Alexander Nikitin, Jannik Kossen, Yarin Gal, and Pekka Marttinen.
\newblock Kernel language entropy: Fine-grained uncertainty quantification for
  llms from semantic similarities, 2024.
\newblock URL \url{https://arxiv.org/abs/2405.20003}.

\bibitem[OpenAI(2024)]{gpt4omini}
OpenAI.
\newblock Hello gpt-4o, 2024.
\newblock URL \url{https://openai.com/index/hello-gpt-4o/}.

\bibitem[OpenAI et~al.(2024)OpenAI, Achiam, Adler, Agarwal, Ahmad, Akkaya,
  Aleman, Almeida, Altenschmidt, Altman, Anadkat, Avila, Babuschkin, Balaji,
  Balcom, Baltescu, Bao, Bavarian, Belgum, Bello, Berdine, Bernadett-Shapiro,
  Berner, Bogdonoff, Boiko, Boyd, Brakman, Brockman, Brooks, Brundage, Button,
  Cai, Campbell, Cann, Carey, Carlson, Carmichael, Chan, Chang, Chantzis, Chen,
  Chen, Chen, Chen, Chen, Chess, Cho, Chu, Chung, Cummings, Currier, Dai,
  Decareaux, Degry, Deutsch, Deville, Dhar, Dohan, Dowling, Dunning, Ecoffet,
  Eleti, Eloundou, Farhi, Fedus, Felix, Fishman, Forte, Fulford, Gao, Georges,
  Gibson, Goel, Gogineni, Goh, Gontijo-Lopes, Gordon, Grafstein, Gray, Greene,
  Gross, Gu, Guo, Hallacy, Han, Harris, He, Heaton, Heidecke, Hesse, Hickey,
  Hickey, Hoeschele, Houghton, Hsu, Hu, Hu, Huizinga, Jain, Jain, Jang, Jiang,
  Jiang, Jin, Jin, Jomoto, Jonn, Jun, Kaftan, Łukasz Kaiser, Kamali,
  Kanitscheider, Keskar, Khan, Kilpatrick, Kim, Kim, Kim, Kirchner, Kiros,
  Knight, Kokotajlo, Łukasz Kondraciuk, Kondrich, Konstantinidis, Kosic,
  Krueger, Kuo, Lampe, Lan, Lee, Leike, Leung, Levy, Li, Lim, Lin, Lin, Litwin,
  Lopez, Lowe, Lue, Makanju, Malfacini, Manning, Markov, Markovski, Martin,
  Mayer, Mayne, McGrew, McKinney, McLeavey, McMillan, McNeil, Medina, Mehta,
  Menick, Metz, Mishchenko, Mishkin, Monaco, Morikawa, Mossing, Mu, Murati,
  Murk, Mély, Nair, Nakano, Nayak, Neelakantan, Ngo, Noh, Ouyang, O'Keefe,
  Pachocki, Paino, Palermo, Pantuliano, Parascandolo, Parish, Parparita,
  Passos, Pavlov, Peng, Perelman, de~Avila Belbute~Peres, Petrov,
  de~Oliveira~Pinto, Michael, Pokorny, Pokrass, Pong, Powell, Power, Power,
  Proehl, Puri, Radford, Rae, Ramesh, Raymond, Real, Rimbach, Ross, Rotsted,
  Roussez, Ryder, Saltarelli, Sanders, Santurkar, Sastry, Schmidt, Schnurr,
  Schulman, Selsam, Sheppard, Sherbakov, Shieh, Shoker, Shyam, Sidor, Sigler,
  Simens, Sitkin, Slama, Sohl, Sokolowsky, Song, Staudacher, Such, Summers,
  Sutskever, Tang, Tezak, Thompson, Tillet, Tootoonchian, Tseng, Tuggle,
  Turley, Tworek, Uribe, Vallone, Vijayvergiya, Voss, Wainwright, Wang, Wang,
  Wang, Ward, Wei, Weinmann, Welihinda, Welinder, Weng, Weng, Wiethoff,
  Willner, Winter, Wolrich, Wong, Workman, Wu, Wu, Wu, Xiao, Xu, Yoo, Yu, Yuan,
  Zaremba, Zellers, Zhang, Zhang, Zhao, Zheng, Zhuang, Zhuk, and Zoph]{gpt4}
OpenAI, Josh Achiam, Steven Adler, Sandhini Agarwal, Lama Ahmad, Ilge Akkaya,
  Florencia~Leoni Aleman, Diogo Almeida, Janko Altenschmidt, Sam Altman,
  Shyamal Anadkat, Red Avila, Igor Babuschkin, Suchir Balaji, Valerie Balcom,
  Paul Baltescu, Haiming Bao, Mohammad Bavarian, Jeff Belgum, Irwan Bello, Jake
  Berdine, Gabriel Bernadett-Shapiro, Christopher Berner, Lenny Bogdonoff, Oleg
  Boiko, Madelaine Boyd, Anna-Luisa Brakman, Greg Brockman, Tim Brooks, Miles
  Brundage, Kevin Button, Trevor Cai, Rosie Campbell, Andrew Cann, Brittany
  Carey, Chelsea Carlson, Rory Carmichael, Brooke Chan, Che Chang, Fotis
  Chantzis, Derek Chen, Sully Chen, Ruby Chen, Jason Chen, Mark Chen, Ben
  Chess, Chester Cho, Casey Chu, Hyung~Won Chung, Dave Cummings, Jeremiah
  Currier, Yunxing Dai, Cory Decareaux, Thomas Degry, Noah Deutsch, Damien
  Deville, Arka Dhar, David Dohan, Steve Dowling, Sheila Dunning, Adrien
  Ecoffet, Atty Eleti, Tyna Eloundou, David Farhi, Liam Fedus, Niko Felix,
  Simón~Posada Fishman, Juston Forte, Isabella Fulford, Leo Gao, Elie Georges,
  Christian Gibson, Vik Goel, Tarun Gogineni, Gabriel Goh, Rapha Gontijo-Lopes,
  Jonathan Gordon, Morgan Grafstein, Scott Gray, Ryan Greene, Joshua Gross,
  Shixiang~Shane Gu, Yufei Guo, Chris Hallacy, Jesse Han, Jeff Harris, Yuchen
  He, Mike Heaton, Johannes Heidecke, Chris Hesse, Alan Hickey, Wade Hickey,
  Peter Hoeschele, Brandon Houghton, Kenny Hsu, Shengli Hu, Xin Hu, Joost
  Huizinga, Shantanu Jain, Shawn Jain, Joanne Jang, Angela Jiang, Roger Jiang,
  Haozhun Jin, Denny Jin, Shino Jomoto, Billie Jonn, Heewoo Jun, Tomer Kaftan,
  Łukasz Kaiser, Ali Kamali, Ingmar Kanitscheider, Nitish~Shirish Keskar,
  Tabarak Khan, Logan Kilpatrick, Jong~Wook Kim, Christina Kim, Yongjik Kim,
  Jan~Hendrik Kirchner, Jamie Kiros, Matt Knight, Daniel Kokotajlo, Łukasz
  Kondraciuk, Andrew Kondrich, Aris Konstantinidis, Kyle Kosic, Gretchen
  Krueger, Vishal Kuo, Michael Lampe, Ikai Lan, Teddy Lee, Jan Leike, Jade
  Leung, Daniel Levy, Chak~Ming Li, Rachel Lim, Molly Lin, Stephanie Lin,
  Mateusz Litwin, Theresa Lopez, Ryan Lowe, Patricia Lue, Anna Makanju, Kim
  Malfacini, Sam Manning, Todor Markov, Yaniv Markovski, Bianca Martin, Katie
  Mayer, Andrew Mayne, Bob McGrew, Scott~Mayer McKinney, Christine McLeavey,
  Paul McMillan, Jake McNeil, David Medina, Aalok Mehta, Jacob Menick, Luke
  Metz, Andrey Mishchenko, Pamela Mishkin, Vinnie Monaco, Evan Morikawa, Daniel
  Mossing, Tong Mu, Mira Murati, Oleg Murk, David Mély, Ashvin Nair, Reiichiro
  Nakano, Rajeev Nayak, Arvind Neelakantan, Richard Ngo, Hyeonwoo Noh, Long
  Ouyang, Cullen O'Keefe, Jakub Pachocki, Alex Paino, Joe Palermo, Ashley
  Pantuliano, Giambattista Parascandolo, Joel Parish, Emy Parparita, Alex
  Passos, Mikhail Pavlov, Andrew Peng, Adam Perelman, Filipe de~Avila
  Belbute~Peres, Michael Petrov, Henrique~Ponde de~Oliveira~Pinto, Michael,
  Pokorny, Michelle Pokrass, Vitchyr~H. Pong, Tolly Powell, Alethea Power,
  Boris Power, Elizabeth Proehl, Raul Puri, Alec Radford, Jack Rae, Aditya
  Ramesh, Cameron Raymond, Francis Real, Kendra Rimbach, Carl Ross, Bob
  Rotsted, Henri Roussez, Nick Ryder, Mario Saltarelli, Ted Sanders, Shibani
  Santurkar, Girish Sastry, Heather Schmidt, David Schnurr, John Schulman,
  Daniel Selsam, Kyla Sheppard, Toki Sherbakov, Jessica Shieh, Sarah Shoker,
  Pranav Shyam, Szymon Sidor, Eric Sigler, Maddie Simens, Jordan Sitkin,
  Katarina Slama, Ian Sohl, Benjamin Sokolowsky, Yang Song, Natalie Staudacher,
  Felipe~Petroski Such, Natalie Summers, Ilya Sutskever, Jie Tang, Nikolas
  Tezak, Madeleine~B. Thompson, Phil Tillet, Amin Tootoonchian, Elizabeth
  Tseng, Preston Tuggle, Nick Turley, Jerry Tworek, Juan Felipe~Cerón Uribe,
  Andrea Vallone, Arun Vijayvergiya, Chelsea Voss, Carroll Wainwright,
  Justin~Jay Wang, Alvin Wang, Ben Wang, Jonathan Ward, Jason Wei, CJ~Weinmann,
  Akila Welihinda, Peter Welinder, Jiayi Weng, Lilian Weng, Matt Wiethoff, Dave
  Willner, Clemens Winter, Samuel Wolrich, Hannah Wong, Lauren Workman, Sherwin
  Wu, Jeff Wu, Michael Wu, Kai Xiao, Tao Xu, Sarah Yoo, Kevin Yu, Qiming Yuan,
  Wojciech Zaremba, Rowan Zellers, Chong Zhang, Marvin Zhang, Shengjia Zhao,
  Tianhao Zheng, Juntang Zhuang, William Zhuk, and Barret Zoph.
\newblock Gpt-4 technical report, 2024.
\newblock URL \url{https://arxiv.org/abs/2303.08774}.

\bibitem[Ouyang et~al.(2022)Ouyang, Wu, Jiang, Almeida, Wainwright, Mishkin,
  Zhang, Agarwal, Slama, Ray, Schulman, Hilton, Kelton, Miller, Simens, Askell,
  Welinder, Christiano, Leike, and Lowe]{ouyang2022training}
Long Ouyang, Jeff Wu, Xu~Jiang, Diogo Almeida, Carroll~L. Wainwright, Pamela
  Mishkin, Chong Zhang, Sandhini Agarwal, Katarina Slama, Alex Ray, John
  Schulman, Jacob Hilton, Fraser Kelton, Luke Miller, Maddie Simens, Amanda
  Askell, Peter Welinder, Paul Christiano, Jan Leike, and Ryan Lowe.
\newblock Training language models to follow instructions with human feedback,
  2022.

\bibitem[Rigollet \& Tong(2011)Rigollet and Tong]{rigollet2011neyman}
Philippe Rigollet and Xin Tong.
\newblock Neyman-pearson classification, convexity and stochastic constraints.
\newblock \emph{Journal of machine learning research}, 2011.

\bibitem[Scott \& Nowak(2005)Scott and Nowak]{scott2005neyman}
Clayton Scott and Robert Nowak.
\newblock A neyman-pearson approach to statistical learning.
\newblock \emph{IEEE Transactions on Information Theory}, 51\penalty0
  (11):\penalty0 3806--3819, 2005.

\bibitem[Skorski(2023)]{skorski2023bernstein}
Maciej Skorski.
\newblock Bernstein-type bounds for beta distribution.
\newblock \emph{Modern Stochastics: Theory and Applications}, 10\penalty0
  (2):\penalty0 211--228, 2023.

\bibitem[Thorne et~al.(2018{\natexlab{a}})Thorne, Vlachos, Christodoulopoulos,
  and Mittal]{thorne2018fever}
James Thorne, Andreas Vlachos, Christos Christodoulopoulos, and Arpit Mittal.
\newblock Fever: a large-scale dataset for fact extraction and verification,
  2018{\natexlab{a}}.

\bibitem[Thorne et~al.(2018{\natexlab{b}})Thorne, Vlachos, Cocarascu,
  Christodoulopoulos, and Mittal]{thorne2018factextractionverificationfever}
James Thorne, Andreas Vlachos, Oana Cocarascu, Christos Christodoulopoulos, and
  Arpit Mittal.
\newblock The fact extraction and verification (fever) shared task,
  2018{\natexlab{b}}.
\newblock URL \url{https://arxiv.org/abs/1811.10971}.

\bibitem[Tong(2013)]{JMLR:v14:tong13a}
Xin Tong.
\newblock A plug-in approach to neyman-pearson classification.
\newblock \emph{Journal of Machine Learning Research}, 14\penalty0
  (92):\penalty0 3011--3040, 2013.
\newblock URL \url{http://jmlr.org/papers/v14/tong13a.html}.

\bibitem[Tong et~al.(2018)Tong, Feng, and Li]{tong2018neyman}
Xin Tong, Yang Feng, and Jingyi~Jessica Li.
\newblock Neyman-pearson classification algorithms and np receiver operating
  characteristics.
\newblock \emph{Science advances}, 4\penalty0 (2):\penalty0 eaao1659, 2018.

\bibitem[Touvron et~al.(2023)Touvron, Lavril, Izacard, Martinet, Lachaux,
  Lacroix, Rozière, Goyal, Hambro, Azhar, Rodriguez, Joulin, Grave, and
  Lample]{touvron2023llama}
Hugo Touvron, Thibaut Lavril, Gautier Izacard, Xavier Martinet, Marie-Anne
  Lachaux, Timothée Lacroix, Baptiste Rozière, Naman Goyal, Eric Hambro,
  Faisal Azhar, Aurelien Rodriguez, Armand Joulin, Edouard Grave, and Guillaume
  Lample.
\newblock Llama: Open and efficient foundation language models, 2023.

\bibitem[Varshney et~al.(2023)Varshney, Yao, Zhang, Chen, and
  Yu]{varshney2023stitch}
Neeraj Varshney, Wenlin Yao, Hongming Zhang, Jianshu Chen, and Dong Yu.
\newblock A stitch in time saves nine: Detecting and mitigating hallucinations
  of llms by validating low-confidence generation, 2023.

\bibitem[Wang et~al.(2023)Wang, Liu, Yue, Tang, Zhang, Jiayang, Yao, Gao, Hu,
  Qi, Wang, Yang, Wang, Xie, Zhang, and Zhang]{wang2023survey}
Cunxiang Wang, Xiaoze Liu, Yuanhao Yue, Xiangru Tang, Tianhang Zhang, Cheng
  Jiayang, Yunzhi Yao, Wenyang Gao, Xuming Hu, Zehan Qi, Yidong Wang, Linyi
  Yang, Jindong Wang, Xing Xie, Zheng Zhang, and Yue Zhang.
\newblock Survey on factuality in large language models: Knowledge, retrieval
  and domain-specificity, 2023.
\newblock URL \url{https://arxiv.org/abs/2310.07521}.

\bibitem[Xiong et~al.(2024)Xiong, Hu, Lu, Li, Fu, He, and
  Hooi]{xiong2024llmsexpressuncertaintyempirical}
Miao Xiong, Zhiyuan Hu, Xinyang Lu, Yifei Li, Jie Fu, Junxian He, and Bryan
  Hooi.
\newblock Can llms express their uncertainty? an empirical evaluation of
  confidence elicitation in llms, 2024.
\newblock URL \url{https://arxiv.org/abs/2306.13063}.

\bibitem[Xu et~al.(2024)Xu, Wu, Diao, Liu, Wang, Chen, and Gao]{xu2024sayself}
Tianyang Xu, Shujin Wu, Shizhe Diao, Xiaoze Liu, Xingyao Wang, Yangyi Chen, and
  Jing Gao.
\newblock Sayself: Teaching llms to express confidence with self-reflective
  rationales.
\newblock \emph{arXiv preprint arXiv:2405.20974}, 2024.

\bibitem[Yang et~al.(2018)Yang, Qi, Zhang, Bengio, Cohen, Salakhutdinov, and
  Manning]{yang2018hotpotqa}
Zhilin Yang, Peng Qi, Saizheng Zhang, Yoshua Bengio, William~W. Cohen, Ruslan
  Salakhutdinov, and Christopher~D. Manning.
\newblock Hotpotqa: A dataset for diverse, explainable multi-hop question
  answering, 2018.

\bibitem[Zhang et~al.(2023)Zhang, Diao, Lin, Fung, Lian, Wang, Chen, Ji, and
  Zhang]{zhang2023r}
Hanning Zhang, Shizhe Diao, Yong Lin, Yi~R Fung, Qing Lian, Xingyao Wang,
  Yangyi Chen, Heng Ji, and Tong Zhang.
\newblock R-tuning: Teaching large language models to refuse unknown questions.
\newblock \emph{arXiv preprint arXiv:2311.09677}, 2023.

\end{thebibliography}
\bibliographystyle{iclr2025_conference}

\newpage
\appendix
\section{Derivation and Proof}
\label{sec:proof}
Assume that when $(q,M(q))\sim P_0$, $\hat\eta(q,M(q))$ has CDF $F$. We denote the $1-\alpha$ quantile of $F$ as $F^{-1}(1-\alpha)=\inf\{x\in\R\mid F(x)\ge1-\alpha\}$. Then we can show that:

\begin{equation}\label{eq:proof}
    \Prob_\D(\Prob_{(q,M(q))\sim P_0}(\hat\eta(q,M(q))>T_{(k)})>\alpha) = 
    \Prob_\D(T_{(k)}< F^{-1}(1-\alpha)) 
\end{equation}

Considering the property of the order statistics, we have that

\begin{align}
    \Prob_\D(T_{(k)} < F^{-1}(1-\alpha)) &= \Prob_\D\left( \sum_{i=1}^{n_0} \mathbb{I}\left(\hat\eta(q_i^{(0)}, M(q_i^{(0)})) < F^{-1}(1-\alpha)\right) \geq k \right) \\
    &\le \sum_{j=k}^{n_0} {n_0 \choose j} (1-\alpha)^j \alpha^{n-j} \overset{\triangle}{=} v(k)
\end{align}

where \( \mathbb{I}(\cdot) \) is the indicator function, defined as:

\[
\mathbb{I}(\hat\eta(q_i^{(0)},M(q_i^{(0)})) < Q_{\alpha}) =
\begin{cases} 
1 & \text{if } \hat\eta(q_i^{(0)},M(q_i^{(0)})) > Q_{\alpha} \\
0 & \text{otherwise}
\end{cases}
\]

\begin{Proof}[Proof of Theorem \ref{thm:type1}]
    Theorem \ref{thm:type1} follows from the definition of $\hat k$.
\end{Proof}

\begin{Lemma}[Theorem 1 in \cite{skorski2023bernstein}]\label{lem:beta}
    Suppose $X_1,\ldots,X_n$ are i.i.d. continuous random variables with CDF $F$, denote 
    \[\epsilon_k=\frac{4(n-2k+1)}{3(n+1)(n+3)}\log\frac{2}{\delta}\vee0+\sqrt{\frac{2k(n-k+1)}{(n+1)^2(n+2)}\log\frac{2}{\delta}},\]
    then
    \[\Prob\bigg(\abs{F(X_{(k)})-\frac{k}{n+1}}\le\epsilon_k\bigg)\ge 1-\delta.\]
\end{Lemma}

\begin{Proof}[Proof of Theorem \ref{thm:power}]

    1) Firstly, we show that $\gR_0(\hat f_\alpha)$ is not much smaller than $\alpha$. To see this, since we assume $\hat\eta(q,M(q))$ with $(q,M(q))\sim P_0$ is a continuous random variable, it follows from the definition of $\hat k$ that
    \[\Prob(F(T_{(\hat k-1)})< 1-\alpha)=\Prob\big(\Prob_{(q,M(q))\sim P_0}(\hat\eta(q,M(q))>T_{(\hat k-1)})>\alpha\big)>\delta.\]
    Here we only consider the case where $\hat k>1$, as will be shown in Equation \eqref{eq:khat_lower_bound}, it holds as long as $n_0$ is not too small. Since $\hat k$ is deterministic given $n_0$, it follows from Lemma \ref{lem:beta} that
    \[\Prob\bigg(F(T_{(\hat k-1)})\le\frac{\hat k-1}{n_0+1}-\epsilon_{\hat k-1}\bigg)\le\delta.\]
    Therefore we have
    \[\frac{\hat k-1}{n_0+1}-\epsilon_{\hat k-1}<1-\alpha.\]
    Denote the event $E_1$ as
    \[E_1=\bigg\{F(T_{(\hat k)})\le\frac{\hat k}{n_0+1}+\epsilon_{\hat k}\bigg\},\]
    it follows from Lemma \ref{lem:beta} that $\Prob(E_1)\ge 1-\delta$. Under $E_1$, we know
    \[F(T_{(\hat k)})\le\frac{\hat k}{n_0+1}+\epsilon_{\hat k}<1-\alpha+\frac{1}{n_0+1}+\epsilon_{\hat k-1}+\epsilon_{\hat k},\]
    which implies
    \[\Prob_{(q,M(q))\sim P_0}(\hat\eta(q,M(q))>T_{(\hat k)})=1-F(T_{(\hat k)})>\alpha-\frac{1}{n_0+1}-\epsilon_{\hat k-1}-\epsilon_{\hat k}.\]
    Similarly, since
    \[\Prob(F(T_{(\hat k)})\ge 1-\alpha)\ge 1-\delta,\quad \Prob\bigg(F(T_{(\hat k)})\le \frac{\hat k}{n_0+1}+\epsilon_{\hat k}\bigg)\ge 1-\delta,\]
    we know
    \[1-\alpha\le\frac{\hat k}{n_0+1}+\epsilon_{\hat k},\]
    which concludes that 
    \begin{equation}\label{eq:khat_lower_bound}
        \hat k\gtrsim (1-\alpha)n_0.
    \end{equation}
    Thus we have
    \[\epsilon_{\hat k-1}+\epsilon_{\hat k}\lesssim \sqrt{\frac{\alpha}{n_0}\log\frac{1}{\delta}},\]
    and
    \begin{equation}\label{eq:error0_lower_bound}
        \Prob_{(q,M(q))\sim P_0}(\hat \eta(q,M(q))>T_{(\hat k)})>\alpha-c\sqrt{\frac{\alpha}{n_0}\log\frac{1}{\delta}}.
    \end{equation}

    2) Secondly, we show $T_{(\hat k)}$ is close to $\tau_\alpha$. Denote $\alpha'=\alpha-c\sqrt{\frac{\alpha}{n_0}\log\frac{1}{\delta}}$, it follows from Equation \eqref{eq:error0_lower_bound} that under $E_1$, we have

    \begin{align*}
    \Prob_{(q,M(q))\sim P_0}\big(\hat\eta(q,M(q)) > T_{(\hat k)}\big) > \alpha' 
    &= \Prob_{(q,M(q))\sim P_0}\big(\eta(q,M(q)) > \tau_{\alpha'}\big) \\
    &\ge \Prob_{(q,M(q))\sim P_0}\big(\hat\eta(q,M(q)) > \tau_{\alpha'} + \epsilon_\eta\big),
\end{align*}

    so
    \[T_{(\hat k)}<\tau_{\alpha'}+\epsilon_\eta.\]
    Denote $E_2$ as
    \[E_2=\{\Prob_{(q,M(q))\sim P_0}(\hat\eta(q,M(q))>T_{(\hat k)})\le\alpha\},\]
    we know $\Prob(E_2)\ge 1-\delta$. Under $E_2$, we have
    \begin{align*}
    \Prob_{(q,M(q))\sim P_0}(\hat\eta(q,M(q))>T_{(\hat k)})\le\alpha&=\Prob_{(q,M(q))\sim P_0}(\eta(q,M(q))>\tau_\alpha)\\ &\le\Prob_{(q,M(q))\sim P_0}(\hat\eta(q,M(q))>\tau_\alpha-\epsilon_\eta),
    \end{align*}
    so
    \[T_{(\hat k)}\ge \tau_\alpha-\epsilon_\eta.\]
    Then we conclude that 
    \[|T_{(\hat k)}-\tau_\alpha|\le\tau_{\alpha'}-\tau_{\alpha}+\epsilon_\eta=\epsilon_\tau.\]

    3) Now we are able to control the excess risk of $\hat f_\alpha$. For any $(q,M(q))$, if we use $Y=0$ (resp. $Y=1$) to denote the model $M$ is uncertain (resp. certain) of $q$, and denote $p_y=\Prob(Y=1)$ to be the marginal distribution for $M$ to be certain, then
    \[\frac{dP_1}{dP_0}(q,M(q))=\frac{\eta(q,M(q))(1-p_y)}{(1-\eta(q,M(q)))p_y},\]
    which is increasing in $\eta(q,M(q))$. Denote $\xi_\alpha=\frac{\tau_\alpha(1-p_y)}{(1-\tau_\alpha)p_y}$, if $|\eta(q,M(q))-\tau_\alpha|\le\epsilon_\tau+\epsilon_\eta$ and $\tau_\alpha+\epsilon_\tau+\epsilon_\eta<1$, then
    \[\bigg|\frac{dP_1}{dP_0}(q,M(q))-\xi_\alpha\bigg|\le\frac{(1-p_y)(\epsilon_\tau+\epsilon_\eta)}{p_y(1-\tau_\alpha-\epsilon_\tau-\epsilon_\eta)^2}.\]
    Then under $E_1\cap E_2$, we can control the excess risk as
    \begin{align*}
        &\gR_1(\hat f_\alpha)-\gR_1(f^*_\alpha)\\
        =&\E_{(q,M(q))\sim P_0}\frac{dP_1}{dP_0}(q,M(q))\bigg(\sI(\hat\eta(q,M(q))\le T_{(\hat k)})-\sI(\eta(q,M(q))\le \tau_\alpha)\bigg)\\
        =&\E_{(q,M(q))\sim P_0}\bigg(\frac{dP_1}{dP_0}(q,M(q))-\xi_\alpha\bigg)\bigg(\sI(\hat\eta(q,M(q))\le T_{(\hat k)})-\sI(\eta(q,M(q))\le \tau_\alpha)\bigg)\\
        &+\xi_\alpha\E_{(q,M(q))\sim P_0}\bigg(\sI(\hat\eta(q,M(q))\le T_{(\hat k)})-\sI(\eta(q,M(q))\le \tau_\alpha)\bigg)\\
        =&\E_{(q,M(q))\sim P_0}\bigg|\frac{dP_1}{dP_0}(q,M(q))-\xi_\alpha\bigg|\bigg|\sI(\hat\eta(q,M(q))\le T_{(\hat k)})-\sI(\eta(q,M(q))\le \tau_\alpha)\bigg|\\&+\xi_\alpha\bigg(\gR_0(f^*_\alpha)-\gR_0(\hat f_\alpha)\bigg)\\
        \le&\E_{(q,M(q))\sim P_0}\bigg|\frac{dP_1}{dP_0}(q,M(q))-\xi_\alpha\bigg|\sI\bigg(|\eta(q,M(q))-\tau_\alpha|\le\epsilon_\tau+\epsilon_\eta\bigg)+c\xi_\alpha\sqrt{\frac{\alpha}{n_0}\log\frac{1}{\delta}}\\
        \lesssim& \frac{(1-p_y)(\epsilon_\tau+\epsilon_\eta)}{p_y(1-\tau_\alpha-\epsilon_\tau-\epsilon_\eta)^2}G_\alpha(\epsilon_\tau+\epsilon_\eta)+\xi_\alpha\sqrt{\frac{\alpha}{n_0}\log\frac{1}{\delta}}.
    \end{align*}
\end{Proof}

\begin{Proof}[Proof of Theorem \ref{thm:type1_shift}]
    It follows from the property of rejection sampling that $\tilde\D_0\mid\II\overset{\rm i.i.d.}{\sim} P_0$, then similar to Theorem \ref{thm:type1}, 
    \[\Prob_\D(\Prob_{(q,M(q))\sim P_0}(\hat f_\alpha(q,M(q))=1)\le\alpha\mid\II)\ge1-\delta.\]
    Taking expectation with respect to $\II$ concludes the result.
\end{Proof}

\section{Implementation Details}
\label{sec:imple}

\begin{figure}[tb!]
    \centering
    \includegraphics[width=0.98\linewidth]{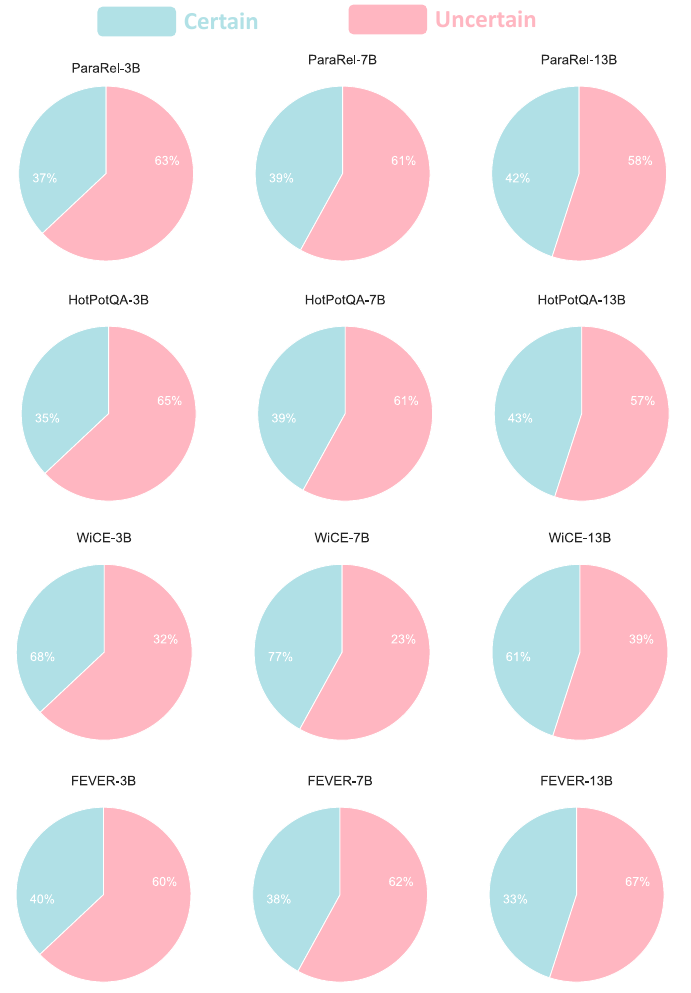}\vspace{20pt}
    \caption{The certain and uncertain data distribution of the originated datasets obtained from supervised identification strategy. The title of each sub-figure consists of the dataset name and the size of the pre-trained model used to evaluate.}
    \label{fig:dataset}
\end{figure}

\subsection{Datasets}
\label{sec:datasets}
We conduct our experiments on four datasets, which are described as follows.
\begin{itemize}[leftmargin=*,label=$\bullet$,noitemsep,partopsep=0pt,topsep=0pt,parsep=0pt]
\item \textbf{ParaRel}~\citep{elazar2021measuring}: This dataset comprises factual knowledge with various prompts and relations initially designed for mask prediction. It is utilized to evaluate the model’s ability to comprehend paraphrased relational facts. To adapt ParaRel for autoregressive models, \citet{zhang2023r} reformatted it into a question-answering format. Duplicate prompts with different templates but identical entities were removed for our question-answering task, resulting in 25,133 prompt-answer pairs across 31 domains. \citet{zhang2023r} divided ParaRel into two subsets: the first 15 domains serve as in-domain data, and the remaining 16 domains as out-of-domain data. The in-domain data is further split equally into training and testing sets.
\item \textbf{HotpotQA}~\citep{yang2018hotpotqa}: It is a question-answering dataset that necessitates complex reasoning across multiple documents. We evaluate the model by providing the relevant context documents and questions to assess its ability to generate correct answers. The development set is used as the testing set for our evaluations.
\item \textbf{WiCE} \citep{kamoi2023wice}: It is a natural language inference (NLI) dataset focused on textual entailment. Each data sample consists of an evidence statement and a claim, and the model must determine whether the evidence supports, partially supports, or does not support the claim. We utilize this dataset as multiple-choice questions with three options for each question.
\item \textbf{FEVER}~\citep{thorne2018fever}: FEVER consists of claims paired with supporting evidence from Wikipedia. Each claim is classified as SUPPORTED, REFUTED, or NOT ENOUGH INFO. This dataset is employed to assess the models' capability to verify the factual accuracy of statements against Wikipedia documents. We utilize FEVER as a multiple-choice NLI task with three options for each question: (A) SUPPORTED, (B) REFUTED, (C) NOT ENOUGH INFO.
\end{itemize}
Detailed information about the original datasets and the data preprocessing procedures can be found in \citet{zhang2023r}. In Figure~\ref{fig:dataset}, we illustrate the distribution of certain and uncertain data within the constructed datasets $D_0, D_1$.

\subsection{Certainty Score Functions}
\label{sec:certain}
We implement our framework with three entropy-based score functions. Details are described as follows.

\begin{itemize}[left=1pt]
    \item \textbf{Semantic Entropy}: Semantic entropy is an entropy which incorporates linguistic invariances created by shared meanings~\cite{semantic}, which is computed by the probability distribution over meanings.
    \begin{equation}
    \label{eq:se1}
    SE(q,M(q)) = - \sum_{c} p(c|q) \log p(c|q) = -\sum_c \bigg(\Big(\sum_{\seq \in c} p(\seq \mid  q)\Big) \log \Big[ \sum_{\seq \in c} p(\seq \mid q) \Big]\bigg)
    \end{equation}
    where $c$ represents possible meaning-class and $p(\seq|q)$ is the probability of the entire answer sequence, that is, the product of the conditional probabilities of new tokens given past tokens. We follow~\cite{semantic} to estimate the expectation of~\ref{eq:se1} given that we cannot have access to all possible $c$. We query $M$ $k$ times and divide the answers into semantic classes $C$ based on semantic equivalence. 
    \begin{equation}
    SE(q,M(q)) \approx -|C|^{-1}\sum_{i=1}^{|C|} \log p(C_i \mid q), \hat\eta(q,M(q)) = -SE(q,M(q)).
    \end{equation}
    Notably, for multiple-choice datasets including WiCE and FEVER, the outputs are among three choices. In this case, we view different tokens as having different semantic meanings, and the semantic entropy is thus reduced to predictive entropy.
    \item \textbf{Kernel Language Entropy}: Kernel language entropy (KLE) is a generalization of semantic entropy~\citep{kle}, providing more detailed uncertainty estimates by considering pairwise semantic dependencies between answers or semantic clusters. It quantifies uncertainty by constructing a semantic kernel from the model's $k$ generated answers and computing its von Neumann entropy. Specifically, for a given input $q$, we generate $k$ responses, build a positive semidefinite semantic kernel $K_\text{sem}$ that captures the semantic relationships among these answers, and then calculate the von Neumann entropy (VNE) of this kernel. The KLE can be defined as:

    \begin{equation}
    \text{KLE}(q,M(q)) = \text{VNE}(K_{sem})=-\Tr[K_\text{sem} \log K_\text{sem}],\ \hat\eta(q,M(q)) = -KLE(q,M(q)).
    \end{equation}

    where, $K_{sem}$ is the semantic kernel which can be implemented from semantic graphs over the LLM outputs.
\end{itemize}

\section{More Experiemnt Results}
\label{sec:moreex}
In this section, we provide more experiment results, including accuracy performance with other significance levels, more Type I error control and certainty distributions.

\subsection{More significance levels.}
\label{sec:morealpha}
\begin{table*}[tb!]
\centering
    \caption{The accuracy performance (\%) of \approach compared to Pretrained models on question-answering and multiple-choice datasets using a significance level of $\alpha = 0.10$. }
    \label{tab:more_acc}
    \resizebox{\linewidth}{!}{
\begin{tabular}{c|c|c|ccc|ccc|c}
\toprule[1.0pt]
Dataset                   & Model                & Pretrained       & $\text{\mini-ve}_5$  & $\text{\mini-ve}_{10}$  & $\text{\mini-ve}_{15}$  & $\text{\mini-se}_5$  & $\text{\mini-se}_{10}$ & $\text{\mini-se}_{15}$ & $\text{\mini-kle}_{15}$ \\
\midrule[0.6pt]
\multirow{3}{*}{ParaRel}  & OpenLLaMA-3B   & \baseline{36.66} & 60.54  & 67.22 & 67.10 & 61.01 & 62.32 & 63.05 & \firstplace{75.51}\\
                          & OpenLLaMA-7B   & \baseline{40.38} & 74.92 & \firstplace{77.84} & 76.89 & 69.00 & 68.78 & 64.97 & 75.36 \\
                          & OpenLLaMA-13B  & \baseline{42.21}  & 77.37 & 75.17 & 79.25 & 68.93 & 68.45 & 68.82 & \firstplace{79.55}\\
\midrule[0.6pt]
\multirow{3}{*}{HotpotQA} & OpenLLaMA-3B   & \baseline{25.72}  & 50.81 & 49.84 & 51.68 & 45.41 & 45.23 & 46.88 & \firstplace{52.70} \\
                          & OpenLLaMA-7B   & \baseline{28.63} & 56.06 & 56.23 & 55.77 & 51.10 & 51.63 & 52.33 & \firstplace{56.73}\\
                          &  LLaMA-13B  & \baseline{30.83}  & 51.49 & 51.45 & 51.61 & 53.42 & 53.12 & 55.38 & 53.34\\
\midrule[0.6pt]
\multirow{3}{*}{WiCE}     & OpenLLaMA-3B   & \baseline{64.72} & 67.65 & 64.40 & \firstplace{76.27} & 61.54 & 64.71 & 64.86 & -- \\
                          & OpenLLaMA-7B   & \baseline{72.96} & 50.00 & 63.77 & 57.32 & 83.33 & \firstplace{85.71} & 74.19 & --\\
                          &  LLaMA-13B  & \baseline{56.89} & 63.33 & 50.00 & 57.14 & 75.00 & 67.44 & \firstplace{77.42} & --\\
\midrule[0.6pt]
\multirow{3}{*}{FEVER}    & OpenLLaMA-3B   & \baseline{39.74} & 60.24 & 53.06 & 52.00  & 80.71 & 82.00 & \firstplace{82.29}  & --\\
                          &     LLaMA-7B & \baseline{35.99} & 43.92 & 43.33 & \firstplace{47.73} & 28.69 & 31.49 & 32.82 & --\\
                          &  LLaMA-13B & \baseline{32.15} & 38.74 & 42.48 & 46.79 & 51.95 & \firstplace{53.01} & 50.92 & --\\
\bottomrule[1.0pt]
\end{tabular}}
\end{table*}

\begin{table*}[tb!]
\centering
    \caption{The Type I error of Fact-Test on question-answering and multiple-choice datasets, with a significance level $\alpha=0.10$.}\vspace{-5pt}
    \label{tab:more_fpr}
    \resizebox{0.9\linewidth}{!}{
\begin{tabular}{c|c|ccc|ccc|c}
\toprule[1.0pt]
Dataset                   & Model                &  $\text{\mini-ve}_5$  & $\text{\mini-ve}_{10}$  & $\text{\mini-ve}_{15}$  & $\text{\mini-se}_5$  & $\text{\mini-se}_{10}$ & $\text{\mini-se}_{15}$ & $\text{\mini-kle}_{15}$  \\
\midrule[0.6pt]
\multirow{3}{*}{ParaRel}  & OpenLLaMA-3B   & 0.0455  & 0.0732 & 0.0783 & 0.0851 & 0.0865 & 0.0905 & 0.0795 \\
                          & OpenLLaMA-7B   & 0.0225 & 0.0240 & 0.0348 & 0.0799 & 0.0886 & 0.0829 & 0.0781\\
                          & OpenLLaMA-13B  & 0.0192  & 0.0226 & 0.0325 & 0.0849 & 0.0706 & 0.0589 & 0.0709\\
\midrule[0.6pt]
\multirow{3}{*}{HotpotQA} & OpenLLaMA-3B   & 0.0276  & 0.0585 &0.0521 & 0.0660 & 0.0678 & 0.0651 & 0.0605 \\
                          & OpenLLaMA-7B   & 0.0295  & 0.0597 & 0.0637 & 0.0631 & 0.0643 & 0.0616 & 0.0590\\
                          &  LLaMA-13B  & 0.0222 & 0.0556 & 0.0675 & 0.0611 & 0.0675 & 0.0503 & 0.0667\\
\midrule[0.6pt]
\multirow{3}{*}{WiCE}     & OpenLLaMA-3B   & 0.0325 & 0.0621 & 0.0414 & 0.0443 & 0.0355 & 0.0325 & --\\
                          & OpenLLaMA-7B   & 0.0694 & 0.0965 & 0.1151 & 0.0347 & 0.0154 & 0.0308  & --\\
                          &  LLaMA-13B  & 0.0266 & 0.0532 & 0.0799 & 0.0169 & 0.0338 & 0.0169  & --\\
\midrule[0.6pt]
\multirow{3}{*}{FEVER}    & OpenLLaMA-3B   & 0.0164 & 0.0418 & 0.0600  & 0.1039 & 0.1053 & 0.1042 & -- \\
                          &     LLaMA-7B & 0.0598  & 0.0617 & 0.0556 & 0.0928 & 0.1027 & 0.1091 & --\\
                          & LLaMA-13B & 0.0172 & 0.0828 & 0.0709  & 0.0944 & 0.1059 & 0.1136 & -- \\
\bottomrule[1.0pt]
\end{tabular}}
\end{table*}

Table~\ref{tab:more_acc} presents the accuracy performance of \approach in comparison with pretrained models at a significance level of $\alpha = 0.10$. Similarly, Table~\ref{tab:more_fpr} displays the corresponding Type I error rates under the same significance level. The results demonstrate that Type I error can still be effectively controlled with the adjusted $\alpha$. Although the accuracies at $\alpha = 0.10$ are somewhat lower than those observed at $\alpha = 0.05$, \approach continues to outperform pre-trained models by a substantial margin while maintaining a lower Type II error rate.

\subsection{Answer Rate Analysis}
\label{sec:answer}
\begin{table*}[tb!]
\centering
    \caption{The Answer Rate and Accuracy Performance (\%) of \approach-t. The number in parenthese is \textcolor{myblue}{Answer Rate}, which means the percentage of willingly answered questions.}\vspace{-5pt}
    \label{tab:answer}
    \resizebox{0.9\linewidth}{!}{
\begin{tabular}{c|c|ccccc}
\toprule[1.0pt]
Dataset                   & Model          &  Finetuned & R-Tuning  & \mini-t $(\alpha=0.15)$ & \mini-t $(\alpha=0.10)$  & \mini-t $(\alpha=0.05)$  \\
\midrule[0.6pt]
\multirow{2}{*}{ParaRel}  & OpenLLaMA-3B   & 61.73 ( \textcolor{myblue}{100\%} ) & 87.42 ( \textcolor{myblue}{37\%} )  & 89.91 ( \textcolor{myblue}{46\%} ) & 92.73 ( \textcolor{myblue}{31\%} )& 94.26 ( \textcolor{myblue}{17\%} )  \\
                          & LLaMA-7B   & 67.73( \textcolor{myblue}{100\%} ) & 89.65 ( \textcolor{myblue}{42\%} ) & 92.76 ( \textcolor{myblue}{47\%} ) & 95.04 ( \textcolor{myblue}{31\%} )& 96.01 ( \textcolor{myblue}{18\%} ) \\
\midrule[0.6pt]
\multirow{2}{*}{FEVER}    & OpenLLaMA-3B   & 65.56 ( \textcolor{myblue}{100\%} ) & 67.19 ( \textcolor{myblue}{11\%} ) & 92.58 ( \textcolor{myblue}{38\%} ) & 94.88 ( \textcolor{myblue}{36\%} ) & 97.82 ( \textcolor{myblue}{33\%} )  \\
                          &  LLaMA-7B & 66.24 ( \textcolor{myblue}{100\%} ) & 66.19 ( \textcolor{myblue}{49\%} )  & 95.41 ( \textcolor{myblue}{28\%} ) & 95.83 ( \textcolor{myblue}{24\%} ) & 96.79 ( \textcolor{myblue}{16\%} ) \\
\bottomrule[1.0pt]
\end{tabular}}
\end{table*}

Table~\ref{tab:answer} presents the answer rate and corresponding accuracy performance (\%) of \approach-t in comparison with baseline methods across multiple datasets and models. The findings demonstrate that \approach-t consistently achieves higher accuracy while effectively managing the answer rate through varying significance levels ($\alpha$). Specifically, \approach-t with $\alpha=0.15$ answers 47\% questions on ParaRel and acheives 92.76\% accuracy, outperforming R-Tuning, which answers 42\% of the questions with an accuracy of 89.65\%. Similarly, \approach-t maintains superior accuracy performance on FEVER compared to baseline models while managing the answer rate through different significance levels.

\subsection{Error Control Analysis}
\label{sec:error}
\begin{figure}[tb!]
\centering
\subfigcapskip=-5pt
   \subfigure[FPR-$\alpha$ curve on HotpotQA]{
    {\includegraphics[width=0.3\linewidth]{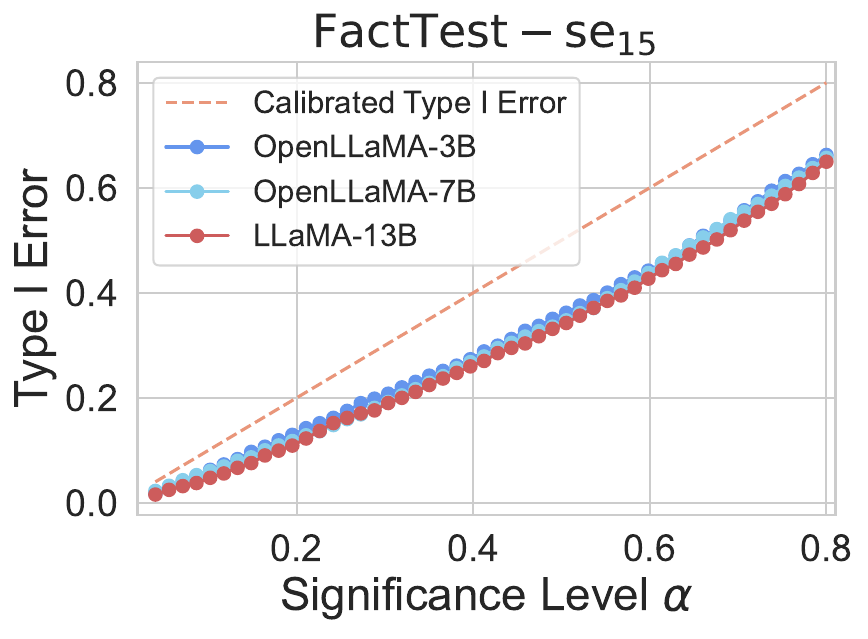}}}\hspace{5pt}
  \subfigure[FNR-$\alpha$ curve on HotpotQA]{
    {\includegraphics[width=0.3\linewidth]{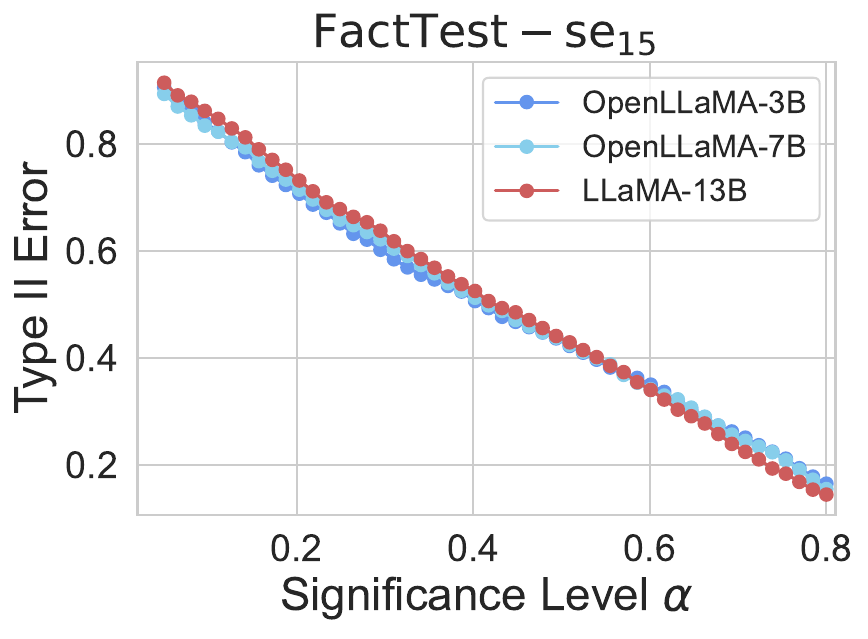}}}

 \caption{The Type I error and Type II error of \approach given different significance levels on HotpotQA using semantic entropy as the certainty function. The legend represents the base pretrained model.}
 \label{fig:hotpot}
\end{figure}

\begin{figure}[tb!]
\centering
\subfigcapskip=-5pt
   \subfigure[\approach-t-3B]{
    {\includegraphics[width=0.3\linewidth]{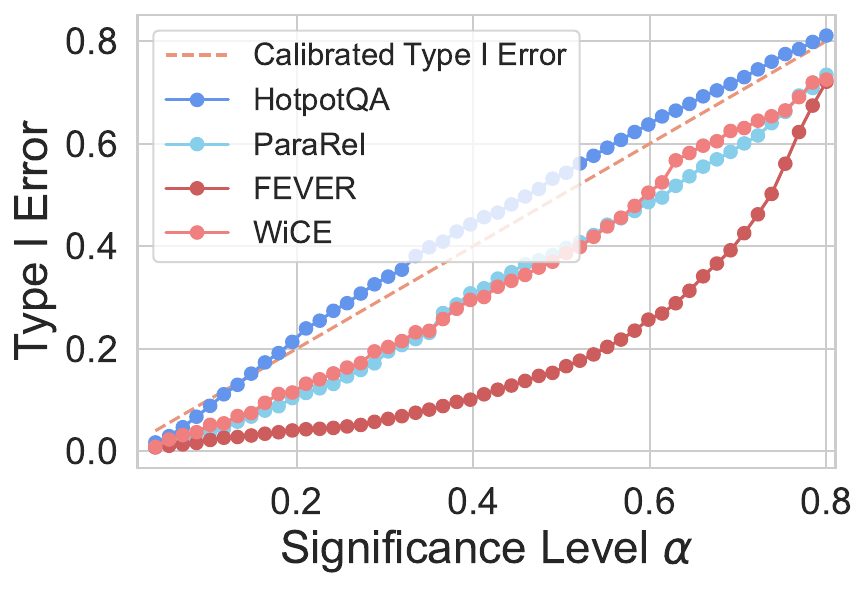}}\label{fig:3b}}\hspace{5pt}
  \subfigure[\approach-t-7B]{
    {\includegraphics[width=0.3\linewidth]{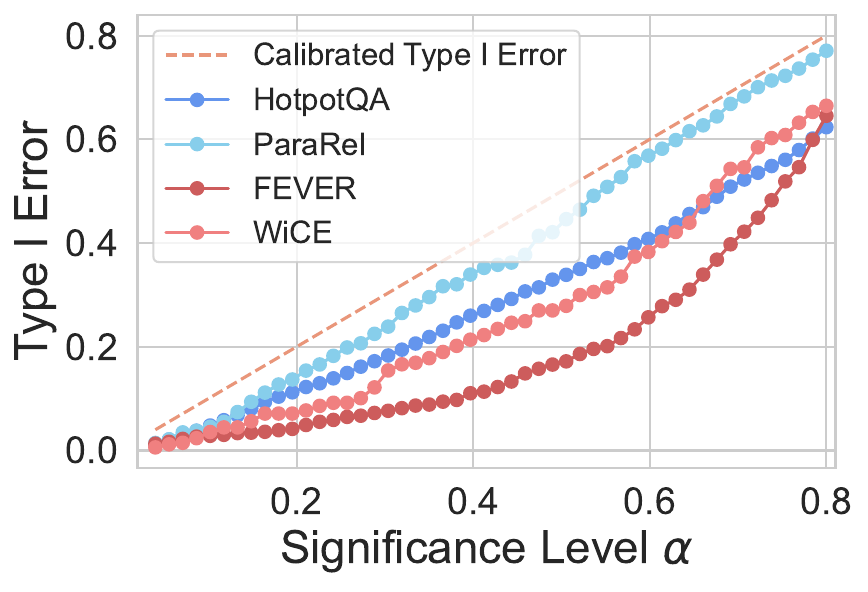}}\label{fig:7b}}

 \caption{The Type I error calibration results of \approach-t given different significance levels using semantic entropy as the certainty function. The legend represents the dataset name.}
 \label{fig:finetune_cali}
\end{figure}

\begin{figure}[tb!]
\centering
\subfigcapskip=-5pt
   \subfigure[ParaRel-3B]{
    {\includegraphics[width=0.3\linewidth]{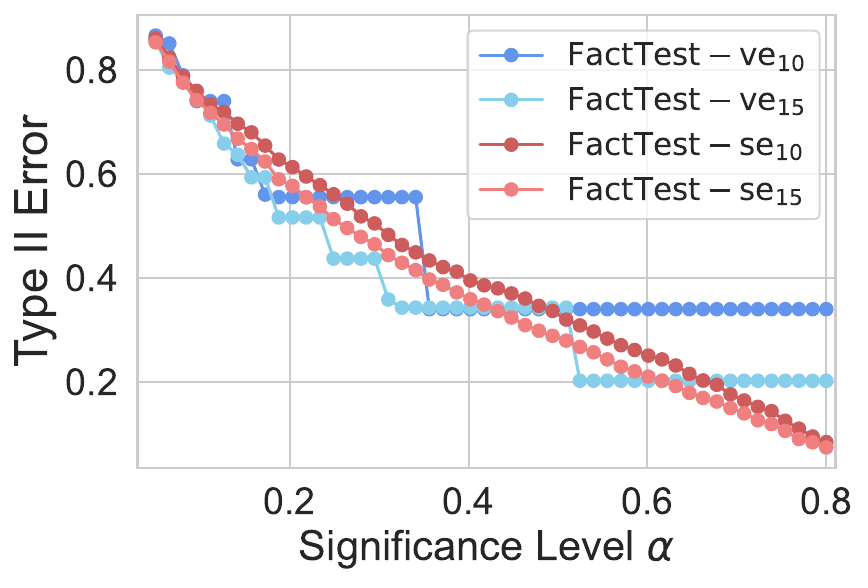}}}\hspace{5pt}
  \subfigure[Hotpot-3B]{
    {\includegraphics[width=0.3\linewidth]{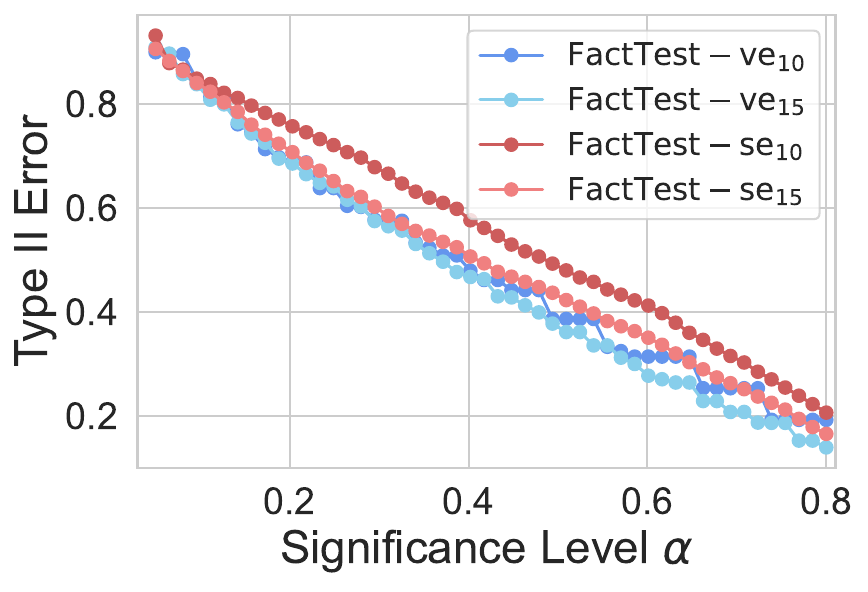}}}\hspace{5pt}
  \subfigure[FEVER-3B]{
    {\includegraphics[width=0.3\linewidth]{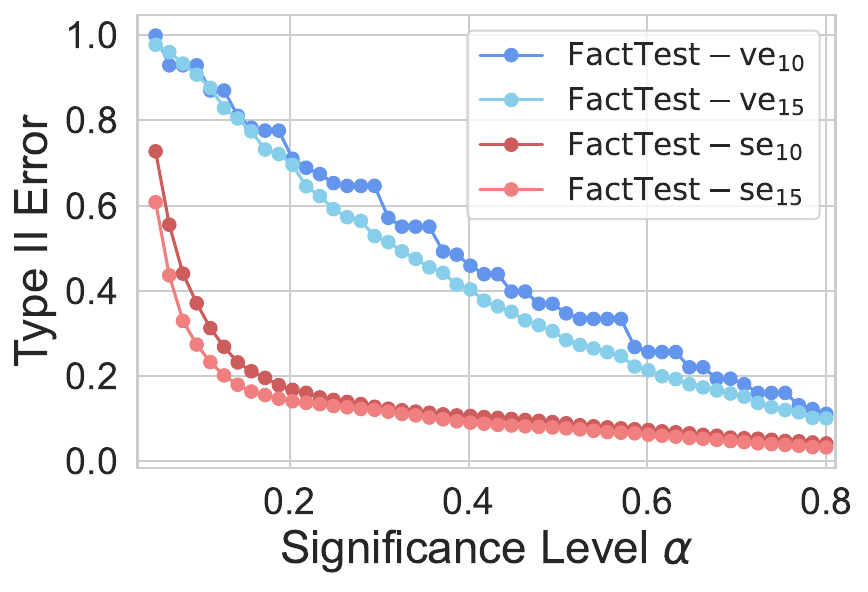}}} 
    \\ \vspace{-10pt}
 \caption{The Type II error simulation of \approach given different significance levels. The caption of each sub-figure consists of the dataset name and the model size. }
  
\label{fig:main_fpr}
\end{figure}


Figure~\ref{fig:hotpot} shows the error control on HotpotQA, demonstrating the ability of \approach to control the Type I error. Figure~\ref{fig:finetune_cali} demonstrates the Type I error calibration results of \approach-t on four datasets, supplementing the experiments in Section~\ref{sec:rtuning}.

\subsection{Certainty Distribution}
\begin{figure}[tb!]
    \centering
    \includegraphics[width=0.98\linewidth]{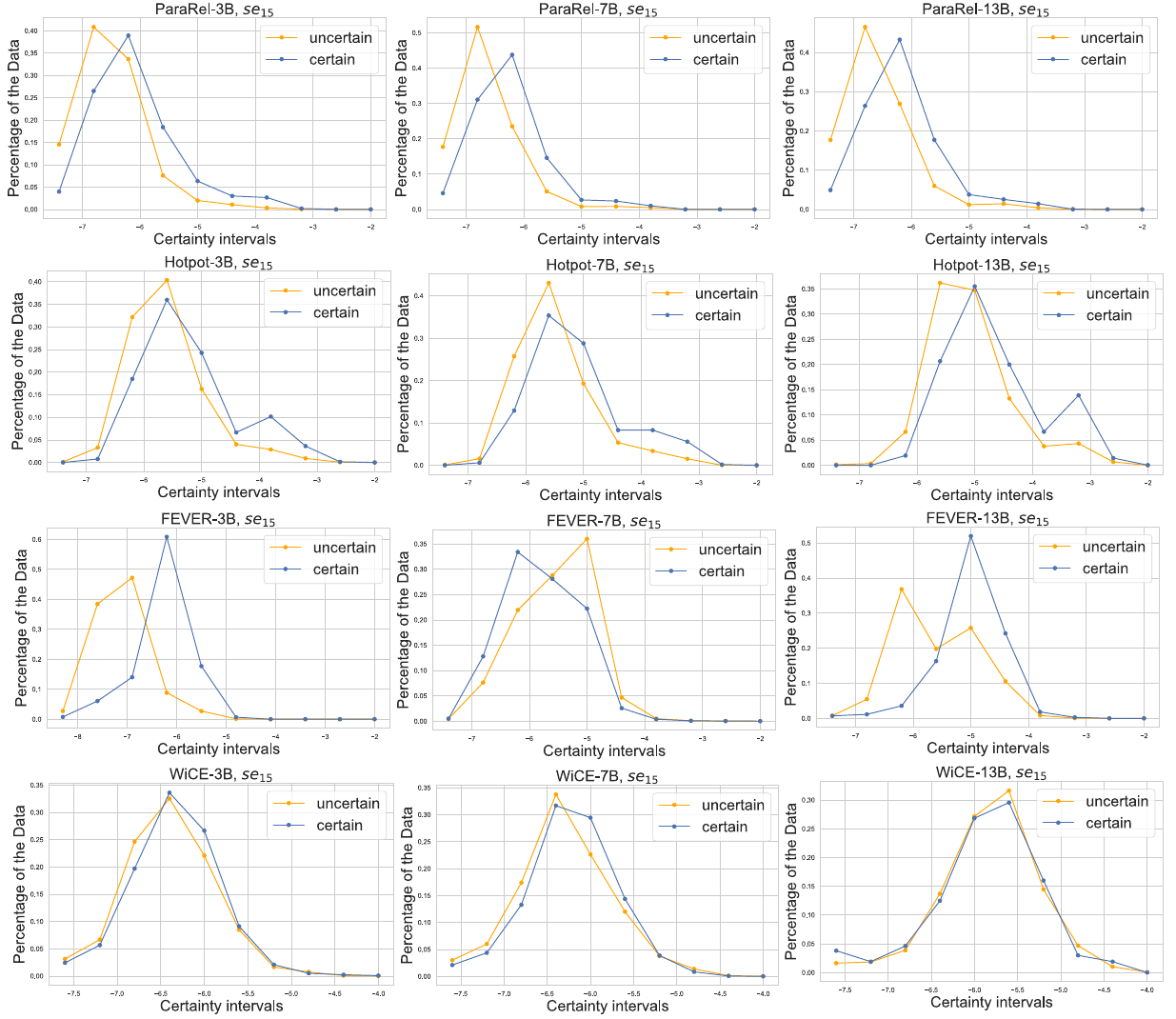}\vspace{-5pt}
    \caption{The certainty distribution of the training datasets on certain set and uncertain set. The title of each sub-figure consists of the dataset name, the size of the pre-trained model used to evaluate, the certainty function and the number of generations.}
    \label{fig:certainty}
\end{figure}

Figure~\ref{fig:certainty} represents the certainty distributions of certain subset and uncertain subset using semantic entropy as the certainty function. The more certain and uncertain distributions can be separated, the better overall performance our framework will be.

\end{document}